\documentclass{article} 
\usepackage{iclr2024_conference,times}
\iclrfinaltrue
\usepackage{booktabs}
\usepackage{subcaption}
\usepackage{graphicx}
\usepackage{placeins}
\usepackage{multirow}
\usepackage{amssymb}
\usepackage{pifont}
\usepackage{algorithm}
\usepackage{algorithmic}
\usepackage{xcolor}
\usepackage{makecell}

\usepackage{amsmath,amsfonts,bm}

\def\eqref#1{equation~\ref{#1}}

\def\1{\bm{1}}

\DeclareMathAlphabet{\mathsfit}{\encodingdefault}{\sfdefault}{m}{sl}
\SetMathAlphabet{\mathsfit}{bold}{\encodingdefault}{\sfdefault}{bx}{n}

\usepackage{chngcntr}
\usepackage{apptools}
\usepackage{footnote}

\usepackage{authblk}

\usepackage{titlesec}
\titlespacing*{\section}{0pt}{3pt}{0pt}

\usepackage{hyperref}
\usepackage{url}

\usepackage{cleveref}
\usepackage{listings}
\definecolor{codegreen}{rgb}{0,0.6,0}
\definecolor{codegray}{rgb}{0.5,0.5,0.5}
\definecolor{codepurple}{rgb}{0.58,0,0.82}
\definecolor{backcolour}{rgb}{0.95,0.95,0.92}
\lstdefinestyle{mystyle}{
    backgroundcolor=\color{backcolour},   
    commentstyle=\color{codegreen},
    numberstyle=\tiny\color{codegray},
    stringstyle=\color{codepurple},
    basicstyle=\ttfamily\footnotesize,
    breakatwhitespace=false,         
    breaklines=true,                 
    captionpos=b,                    
    keepspaces=true,                 
    numbers=left,                    
    numbersep=5pt,                  
    showspaces=false,                
    showstringspaces=false,
    showtabs=false,                  
    tabsize=2
}

\crefname{table}{Tab.}{Tabs.}
\crefname{figure}{Fig.}{Figs.}
\crefname{appendix}{Appendix}{}

\title{From spectra to biophysical insights: end-to-end learning with a biased radiative transfer model}

\author[1]{Yihang She\thanks{ys611@cam.ac.uk}\hspace{2mm}}
\author[2]{Clement Atzberger\thanks{clement@mantle-labs.com}\hspace{2mm}}
\author[1, 2]{Andrew Blake\thanks{ab@ablake.ai}\hspace{2mm}}
\author[1]{Srinivasan Keshav\thanks{sk818@cam.ac.uk}\hspace{2mm}}

\affil[1]{University of Cambridge}
\affil[2]{Mantle Labs}

\begin{document}

\maketitle

\begin{abstract}
Advances in machine learning have boosted the use of Earth observation data for climate change research. Yet, the interpretability of machine-learned representations remains a challenge, particularly in understanding forests' biophysical reactions to climate change. Traditional methods in remote sensing that invert radiative transfer models (RTMs) to retrieve biophysical variables from spectral data often fail to account for biases inherent in the RTM, especially for complex forests. 
We propose to integrate RTMs into an auto-encoder architecture, creating an end-to-end learning approach. Our method not only corrects biases in RTMs but also outperforms traditional techniques for variable retrieval like neural network regression. Furthermore, our framework has potential generally for inverting biased physical models.
The code is available on \url{https://github.com/yihshe/ai-refined-rtm.git}.
\end{abstract}
\section{Introduction}\label{sec:intro}
Over the last few decades, satellite remote sensing has made a vast amount of Earth observation data available \citep{claverie2018harmonized}. 
However, traditional statistical learning approaches in ML are unable to yield interpretable representations \citep{chen2016infogan} which restricts their applications.
In the context of climate change, one significant example concerns understanding the biophysical variables of forests and their response to climate threats \citep{thompson2009forest}, for example: leaf area index, pigmentation, and crown coverage, all of which have direct implications on ecosystem resilience or susceptibility to climate change. 

Physical models, such as Radiative Transfer Models (RTMs) \citep{rosema1992new}, have been developed to simulate canopy-radiation interactions and open pathways for retrieving biophysical variables from satellite spectra. Classical approaches for inverting RTMs include numerical optimization \citep{goel1988models}, look-up tables \citep{combal2003retrieval}, and neural network regressions \citep{gong1999inverting} which is regarded as the state-of-the-art.
Despite decades of research, the complexity of forest structures leads to discrepancies between simulated and measured spectra --- biases that significantly affect the accuracy of estimated biophysical variables using classical approaches \citep{widlowski2013fourth}. Inverting RTMs is challenging also due to the correlations of variables and the ill-posed nature of the inversion problem.

To tackle these challenges, we propose an end-to-end approach to learning biophysical variables from satellite spectra while simultaneously correcting the systematic biases of the RTM. Deviating from traditional methods, we draw inspiration from physics-informed machine learning \citep{lu2020extracting, zerah2022physics, hao2022physics} and disentangled representation learning \citep{eastwood2018framework, locatello2019challenging}. 

Specifically, we use an auto-encoder architecture, training the encoder to predict RTM variables based on input spectra. The RTM is integrated as the decoder, with additional non-linear layers to correct the systematic bias (\cref{fig:models}). 
Once trained, the encoder component can be used independently to extract more accurate biophysical variables from forest spectra, and the bias correction module can be used to generate more realistic forest signatures. 

Our contributions can be summarised as follows:
\begin{itemize}
    \item We have developed an end-to-end learning approach with a complex physical model to retrieve biophysical variables from satellite spectral data.
    \item By incorporating a bias correction layer, we have effectively corrected the bias, resulting in our model's learned variables outperforming state-of-the-art methods for variable retrieval.
    \item Our methodology for inverting the RTM has broader implications for inverting biased physical models.
\end{itemize}

\section{Related Work}
\subsection{Disentangled representation learning}
Disentangled representation learning aims to uncover lower dimensional and semantically meaningful factors of variation from high-dimensional data \citep{locatello2019challenging}. Our approach to inferring biophysical variables by inverting a physical model is partly related to previous work on analysis-by-synthesis for faces: Active Shape Models \citep{cootes1995active} iteratively refine pose, scale, and shape parameters, progressively minimizing the loss between the observed data and the output of a ``Point Distribution Model". However, where those approaches compute each application of the inverse, we seek to learn the inverse function itself, represented as a neural network. 

State-of-the-art methods for unsupervised learning of disentangled representations often use the Variational Autoencoder (VAE) \citep{kingma2013auto} and its variants e.g. $\beta$-VAE \citep{higgins2017beta}. VAE approaches are based on probabilistic modeling \citep{locatello2019challenging} and allow disentangled variables to emerge from learning \citep{kumar2018variational}. 
Alternatively, Generative Adversarial Networks (GAN) \citep{goodfellow2020generative} have also shown success in learning interpretable and disentangled representations for image synthesis, for example InfoGAN \citep{chen2016infogan} and StyleGAN \citep{karras2019style}. However, as generative models,  disentangled variables usually emerge only from learning and they lack an effective inference mechanism~(citep{kumar2018variational}. 

For physical models however, as here, the situation is simplified in that the forward model is deterministic and benefits from having a predetermined set of interpretable variables, namely the physical input parameters to the model.  
Therefore, we use a neural auto-encoder architecture to learn an inverse function that directly predicts disentangled biophysical variables from input spectra.

\subsection{Physics-informed machine learning}
Methods combining physical knowledge with machine learning vary depending on problem context and physical constraint representation \citep{hao2022physics}. Inverse problems represent a popular paradigm within this field, aiming to infer physical parameters from data while satisfying given physical constraints. In these approaches, auto-encoders usually serve as the base of the physics-informed architecture, where a known physical model acts as a fixed decoder and the model learns a complementary encoder as the inverse. For instance, \citet{lu2020extracting} apply a VAE-based architecture to predict latent parameters controlling the dynamics of a system governed by partial differential equations. Similarly, \citet{zerah2022physics} introduce ``pheno-VAE" to extract the phenological parameters from NDVI time series, which also uses VAE-based architecture with a temporal model as the decoder, describing the shape of the time series. While pheno-VAE is one of the few attempts to deal with remote sensing data, the inverse is taken to be simply a double-logistic function, making the inversion process quite straightforward. In contrast, our approach accommodates the more complex physical models often needed to explain remote sensing data, and hence also for analysing such data. 

\subsection{Canopy reflectance models and inversion}
Canopy reflectance models simulate spectral bidirectional forest reflectance given the inputs of important forest characteristics \citep{widlowski2013fourth}. There are four main types of canopy models, namely: geometrical models \citep{li1985geometric}, analytical models e.g. RTMs~\citep{suits1971calculation}, hybrids of geometrical and analytical models \citep{atzberger2000development}, and ray-tracing models \citep{gastellu1996modeling}. Ray-tracing offers accuracy but at the cost of increased computational complexity, which limits its feasibility for model inversion. The model we study in the project is the Invertible Forest Reflectance Model (INFORM) \cite{atzberger2000development}, which is a hybrid combining both geometrical models and RTM. INFORM can simulate a more realistic relationship between canopy and reflectance, while striking a balance between model complexity and computational efficiency, well-suited for inverse problems. 

Classical approaches to invert RTMs include numerical optimizations \citep{goel1988models}, look-up tables \citep{combal2003retrieval} and neural network regression \citep{gong1999inverting, schlerf2006inversion}. However, none of these methods consider the systematic bias that a fixed RTM typically exhibits, owing to its only approximate representation of a given physical system.

We do not claim originality in using an auto-encoder to invert physical models --- this concept has been investigated as above and, more recently, specifically for RTMs~\citep{zerah2023physics}. 
However, we highlight that systematic bias is a critical factor affecting the quality of variables retrieved from RTMs. Our contribution, that we believe to be original, is to combine an auto-encoder with a bias correction network, to significantly improve inverse computation. Furthermore, we introduce a comprehensive framework that seamlessly incorporates complex physical models into an end-to-end learning pipeline, encompassing model conversion, forward pass implementation, and backpropagation. This is applicable quite generally to computing an inverse function for a given fixed physical models with bias correction included. In that way, in addition to obtaining the inverse function, the original model is also bias corrected.
\section{Methods}
\subsection{INFORM and variables to learn}
To test our proposed approach with forestry applications we use the INFORM model \citep{atzberger2000development}. INFORM consists of a group of sub-models to model the reflectance of forest canopies at different levels. Each sub-model has a set of input parameters describing the biophysical and geometric features of the canopy components (\cref{tab:para_list}). We aim to invert INFORM to estimate certain biophysical variables within known ranges (\cref{tab:para_list}). We fix other variables at typical values for the study area. Crown diameter $cd$ and canopy height $h$ are derived from fractional coverage $fc$ using allometric equations \citep{jucker2017allometric}.

The uncertainties in a physical model can be classified into three categories: aleatory, which is inherent to probabilistic models; epistemic, stemming from insufficient knowledge of model parameters; and ontological, resulting from model incompleteness. 
The bias we seek to mitigate arises from the ontological uncertainty of the model. Specifically, the RTM exhibits bias due to incomplete forest modeling, necessitated by simplified assumptions in representing canopy-radiation interactions. Consequently, detecting and correcting such biases is crucial for accurate biophysical variable retrieval.

\begin{table*}[htbp] 
\caption{\textbf{The biophysical variables used by INFORM.} \textit{Variables can be attributed to three hierarchical levels. Seven variables will be inferred directly. \textbf{* Note} that $cd$ will be inferred from $fc$ based on their geometric relationship and $h$ will be inferred from $cd$ using an allometric equation derived from the database compiled by \citet{jucker2017allometric}} (see 
\cref{appx:vars}).}
\label{tab:para_list}
\centering
\resizebox{\textwidth}{!}{
    \begin{tabular}{lcccccc}
    \toprule
    \multirow{2}*{Group} & \multirow{2}*{Variable} & \multirow{2}*{Acronym} & \multirow{2}*{To infer} & \multirow{2}*{Default Value} & \multicolumn{2}{c}{Sample Range} \\
    \cmidrule(lr){6-7}
    ~ & ~ & ~ & ~ & ~ & Min & Max \\
    \midrule
    Background & Soil brightness factor & psoil & \ding{55} & 0.8 & - & - \\
    \midrule
    \multirow{9}*{Leaf Model} & Structure Parameter & N & \ding{51} & - & 1 & 3 \\
    ~ & Chlorophyll A+B & cab & \ding{51} & - & 10 & 80 \\
    ~ & Water Content & cw & \ding{51} & - & 0.001 & 0.02 \\
    ~ & Dry Matter & cm & \ding{51} & - & 0.005 & 0.05 \\
    ~ & Carotenoids & car & \ding{55} & 10 & - & - \\
    ~ & Brown Pigments & cbrown & \ding{55} & 0.25 & - & - \\
    ~ & Anthocyanins & anth & \ding{55} & 2 & - & - \\
    ~ & Proteins & cp & \ding{55} & 0.0015 & - & - \\
    ~ & Cabon-based Constituents & cbc & \ding{55} & 0.01 & - & - \\
    \midrule
    \multirow{7}*{Canopy Model} & Leaf Area Index & LAI & \ding{51} & - & 0.01 & 5 \\
    ~ & Leaf Angle Distribution & typeLIDF & \ding{55} & Beta Distribution & - & - \\
    ~ & Hot Spot Size & hspot & \ding{55} & 0.01 & - & - \\
    ~ & Observation Zenith Angle & tto & \ding{55} & 0 & - & - \\
    ~ & Sun Zenith Angle & tts & \ding{55} & 30 & - & - \\
    ~ & Relative Azimuth Angle & psi & \ding{55} & 0 & - & - \\
    \midrule
    \multirow{4}*{Forest Model} & Undergrowth LAI & LAIu & \ding{51} & - & 0.01 & 1 \\
    ~ & Stem Density & sd & \ding{55} & 500 & - & - \\
    ~ & Fractional Coverage & fc & \ding{51} & - & 0.1 & 1 \\
    ~ & Tree Height & h & \ding{51}* & - & - & - \\
    ~ & Crown Diameter & cd & \ding{51}* & - & - & - \\
    \bottomrule
    \end{tabular}
}
\end{table*}

\subsection{End-to-end learning with INFORM}
To incorporate an RTM such as INFORM into an end-to-end learning pipeline, it must be made fully differentiable. However, like many other RTMs, INFORM is implemented using Numpy arrays and operations --- it is sufficient for the use to create a lookup table or to create a synthetic dataset for neural network regression --- however, it has no definitions on how derivatives should be calculated for each operation. Thus it is necessary to reimplement INFORM using a machine learning framework such as PyTorch, which can track the computational graph and enable backpropagation. Differentiability is crucial for integrating physical models into machine learning workflows. For example, \citet{loper2014opendr} developed a renderer in Chumpy that facilitates differentiable rendering, eliminating the need for manual derivative computations.

However, INFORM is a complicated physical model, and reimplementing it in PyTorch line by line seems challenging. In light of the recent development of large language models, we decide to utilize GPT-4 \citep{openai2023gpt4} to assist in the conversion from Numpy to PyTorch. 
GPT-4 is useful for converting commonly seen operations from NumPy into PyTorch. With its assistance, we have successfully converted 1,742 lines of code across various scripts, from the original implementation, into PyTorch (see \cref{appx:rtm_conversion} for details).

As a physical model, input variables of INFORM carry specific physical meanings and are inherently non-negative in real-world scenarios, as outlined in \cref{tab:para_list}. Providing input variables outside their defined value ranges can result in failure during the forward pass. For instance, the structure parameter 
$N$ acts as a denominator in some calculations, and setting it to 0 will cause computational errors and result in infinite numerical values in the simulated spectra.
To mitigate these issues, we employ a Sigmoid function to learn a scaling factor for each input variable, ensuring the factor ranges between 0 and 1. Subsequently, we rescale these normalized factors to their respective known ranges as specified in \cref{tab:para_list} (see \cref{appx:foward_pass} for more details). This process not only guarantees that the output of INFORM remains meaningful and within expected bounds but also narrows the search space for biophysical parameters. By adjusting input variables to avoid improbable ranges, we enhance the model's reliability in the forward pass and ensure a more effective search for the inversion process. 

The steps so far should have obtained a fully differentiable INFORM. In practice, modelling the canopy-radiation interaction needs exponential, logarithm, and square root functions which can lead to numerically unstable derivative computations. An effective workaround ensures the optimization process bypasses instability points during backpropagation to allow continuous convergence (see \cref{appx:backprop} and \cref{appx:stab_train}). 

These combined measures enable the integration of a complex physical model like INFORM into an end-to-end learning pipeline. Our approach to converting INFORM into PyTorch, including our designs for the forward pass and backpropagation process, could be generalized to incorporate other physical models into machine learning pipelines.

\subsection{Dataset}
The real dataset we are using (denoted by $D_r$) consists of 17962 individual spectra (denoted by $X_r = \{x_{r, 1}, \dots, x_{r, N}\}$) extracted from 1283 individual sites across Austria. Samples of $D_r$ are extracted from each site as a temporal sequence covering April to October 2018 --- both coniferous and deciduous forests, consisting of 12 species. $X_r$ consists of 11 bands used to retrieve biophysical variables.
Collecting ground truth data for biophysical variables is largely impractical. Therefore, the interpretation of results has to rely on plausibility checks --- on variable distributions and temporal evolution patterns. We use available prior knowledge of tree species and temporal data to validate our model's biophysical estimates. For example, $LAI$ inferred from a given site's spectra should exhibit distinct and repeatable temporal trends. 
\begin{table*}
\caption{\textbf{Sentinel-2 bands} to use. \textit{VNIR stands for Visible and Near Infrared. SWIR stands for Short Wave Infrared.}}
\label{tab:s2_bands}
\centering
\resizebox{\textwidth}{!}{
    \begin{tabular}{lccccccccccc}
    \toprule
    Band & B2 & B3 & B4 & B5 & B6 & B7 & B8 & B8a & B9 & B11 & B12 \\
    \midrule
    Resolution & 10 m & 10 m & 10 m & 20 m & 20 m & 20 m & 10 m & 20 m & 60 m & 20 m & 20 m \\
    Central Wavelength & 490 nm & 560 nm & 665 nm & 705 nm & 740 nm & 783 nm & 842 nm & 865 nm & 940 nm & 1610 nm & 2190 nm \\
    Description & Blue & Green & Red & VNIR & VNIR & VNIR & VNIR & VNIR & SWIR & SWIR & SWIR \\
    \bottomrule
    \end{tabular}
}
\end{table*}

\subsection{Inverting the RTM}
\begin{figure}[htbp]
    \centering
    \begin{subfigure}{\textwidth}
        \centering
        \includegraphics[width=1\textwidth]{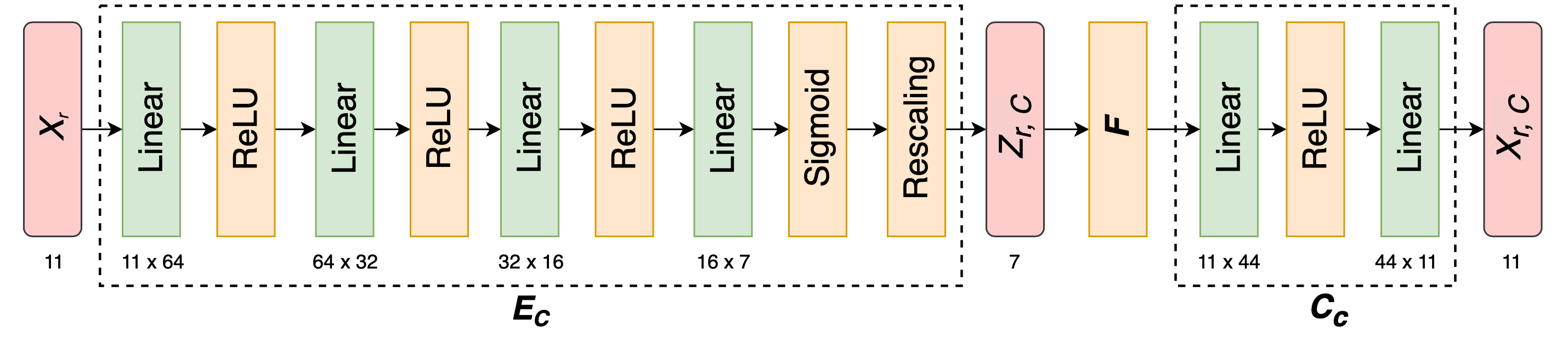}
        \caption{AE\_RTM\_corr}
        \label{fig:model_corr}
    \end{subfigure}
    \begin{subfigure}{\textwidth}
        \centering
        \includegraphics[width=1\textwidth]{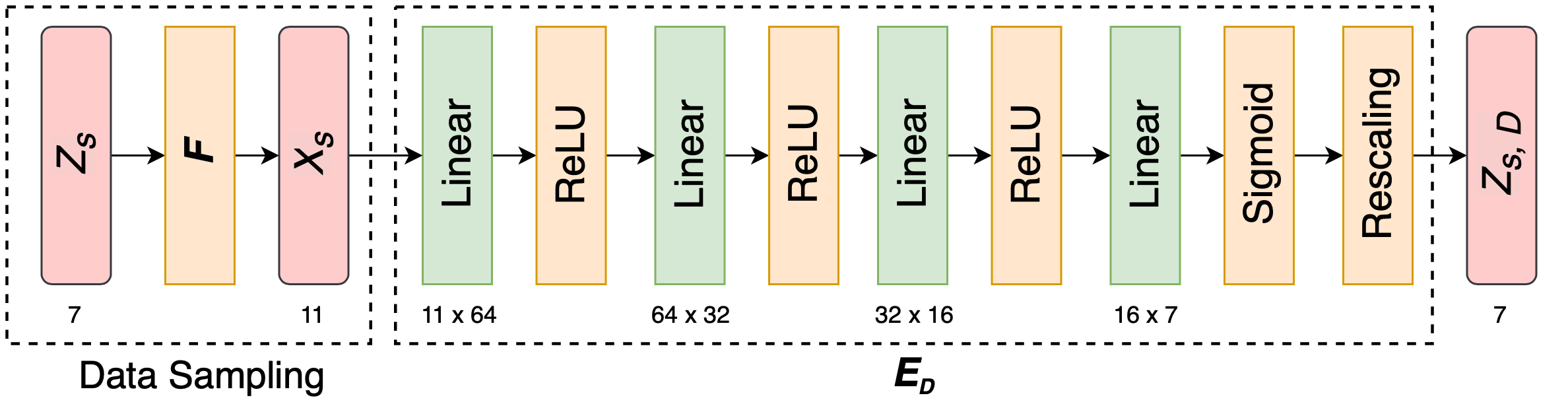}
        \caption{Data sampling and NNRegressor}
        \label{fig:model_nn}
    \end{subfigure}
    \caption{\textbf{Learning the inverse.} \textit{(a) End-to-end learning of the inverse $\mathbf{E_C}$ of RTM $\mathbf{F}$ together with bias correction function $\mathbf{C}_C$. (b) Classical approach --- sampling $D_s$ to train regressive neural network $\mathbf{E_D}$ --- serves as a baseline.}}
    \label{fig:models}
\end{figure}

To learn biophysical variables from  satellite spectra, we use an auto-encoder (AE) architecture extended with bias correction layers to train an encoder that maps measured spectra to biophysical variables. We replace the decoder of the AE with INFORM (\cref{fig:models}), in order to correct the systematic bias of INFORM while refining the learning of biophysical variables (see appendix for training details). 

%
A classical auto-encoder consists of an encoder $\mathbf{E}_{A}$ and a decoder $\mathbf{D}_{A}$ (\cref{eq:vanilla_ae}). 
\begin{equation} \label{eq:vanilla_ae}
    X_{r, A} = \mathbf{D}_{A}(\mathbf{E}_{A}(X_r))
\end{equation}
%
By replacing $\mathbf{D}_{A}$ in \cref{eq:vanilla_ae} with INFORM (denoted by $\mathbf{F}$), we ensure that the encoder (denoted by $\mathbf{E}_B$) will embed $X_r$ as biophysical variables $Z_{r, B}$ (\cref{eq:ae_rtm_encoder}), which are then passed to $\mathbf{F}$ to reconstruct spectra $X_{r, B}$ (\cref{eq:ae_rtm_decoder}).
\begin{equation} \label{eq:ae_rtm_encoder}
    Z_{r, B} = \mathbf{E}_{B}(X_{r})
\end{equation} 
\begin{equation} \label{eq:ae_rtm_decoder}
    X_{r, B} = \mathbf{F}(\mathbf{E}_{B}(X_r))
\end{equation}
%
The final model, AE\_RTM\_corr, has additional non-linear layers $\mathbf{C}_{C}$ (\cref{eq:ae_rtm_corr}) to correct the systematic bias of $\mathbf{F}$. 
\begin{equation} \label{eq:ae_rtm_corr}
    X_{r, C} = \mathbf{C}_{C}(\mathbf{F}(\mathbf{E}_{C}(X_r)))
\end{equation}
%
\subsection{Baseline}\label{methods:baseline}
We have sampled a synthetic dataset $D_s = \{Z_s, X_s\}$ using INFORM, following \cref{tab:para_list}. $D_s$ consists of sampled variables $Z_s$ and synthetic spectra $X_s$. 
Using $D_s$, we train a neural network regression model ``NNRegressor'' replicating the classical approach to inverting the RTM \citep{gong1999inverting}. NNRegressor (model architecture denoted by $\mathbf{E}_D$) is first trained on $D_s$ and then evaluated on the test set of $D_r$ to predict $Z_{r, D}$ given $X_r$. To get the spectral reconstruction $X_{r, D}$, $Z_{r, D}$ will be passed to $\mathbf{F}$.

\subsection{Implementation details}
Both $D_r$ and $D_s$ are further split into the train, validation, and test sets, where samples from the same individual site are placed in the same set to avoid data leakage. All input data are standardized before training. For model training, the batch size is 64. We use the Adam optimizer and set the initial learning rate as 1e-3 and weight decay as 1e-4. The training loss is mean squared error (MSE). The maximum number of epochs is 100, the learning rate is reduced by a factor of 10 after 50 epochs, and training stops if the MSE loss on the validation set does not improve after an additional 10 epochs. We use 11 spectral bands excluding B1 and B10 to allow fair comparison, as these two bands are not available in $D_r$. 
\section{Results}
\subsection{Bias correction}
\begin{figure}[htbp]
    \centering
    \includegraphics[width=1\textwidth]{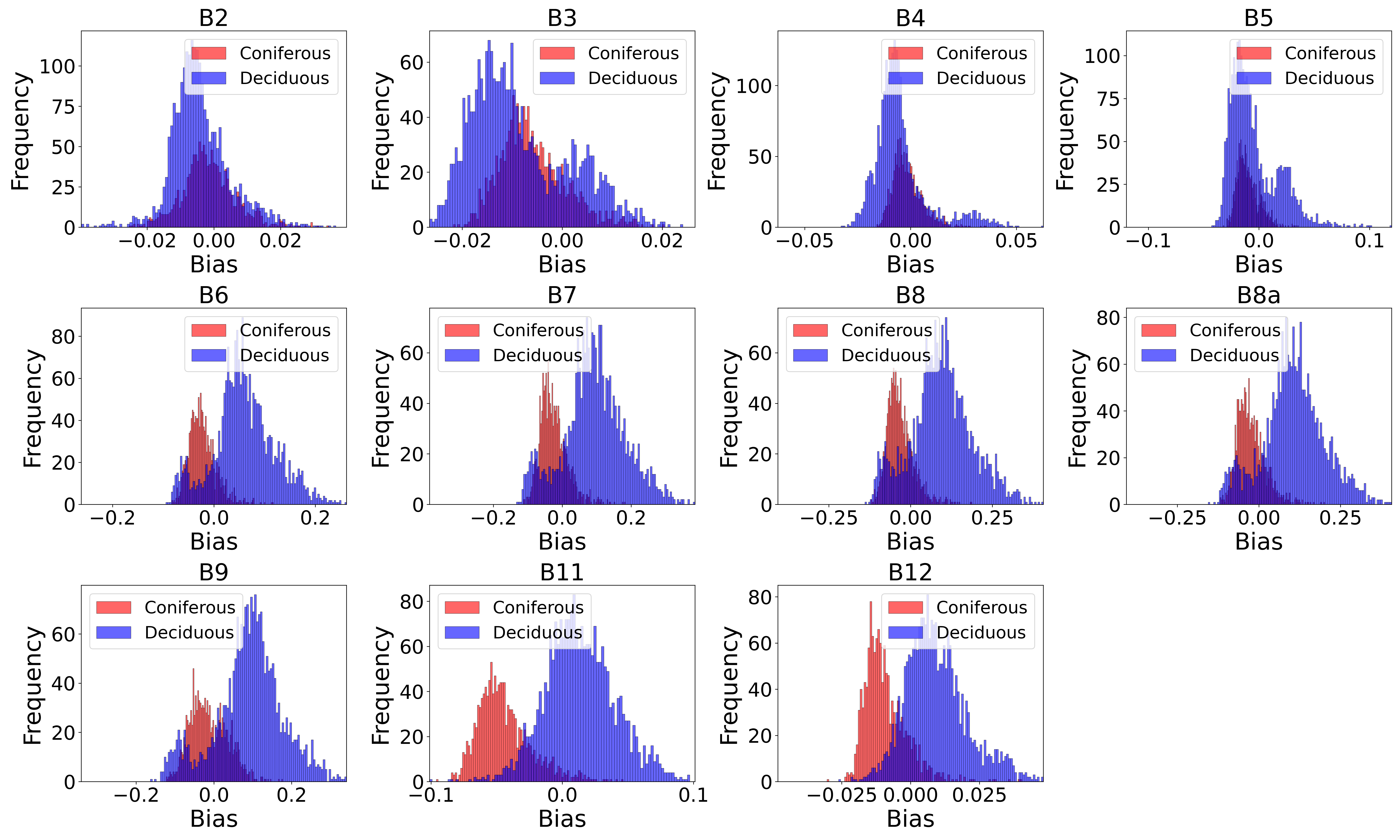}
    \caption{\textbf{Distributions of biases learned by AE\_RTM\_corr}. \textit{The biases are computed by subtracting the corrected spectra from the originally simulated spectra, given the same set of inferred variables. For bands in the near-infrared and short wave ranges, the RTM tends to under-estimate the spectra for deciduous forest but over-estimate for coniferous forest.}}
    \label{fig:hist_bias_corr_coni_v_deci}
\end{figure}

\begin{figure}[htbp]
    \centering
    \begin{subfigure}{\textwidth}
        \centering
        \includegraphics[width=1\textwidth]{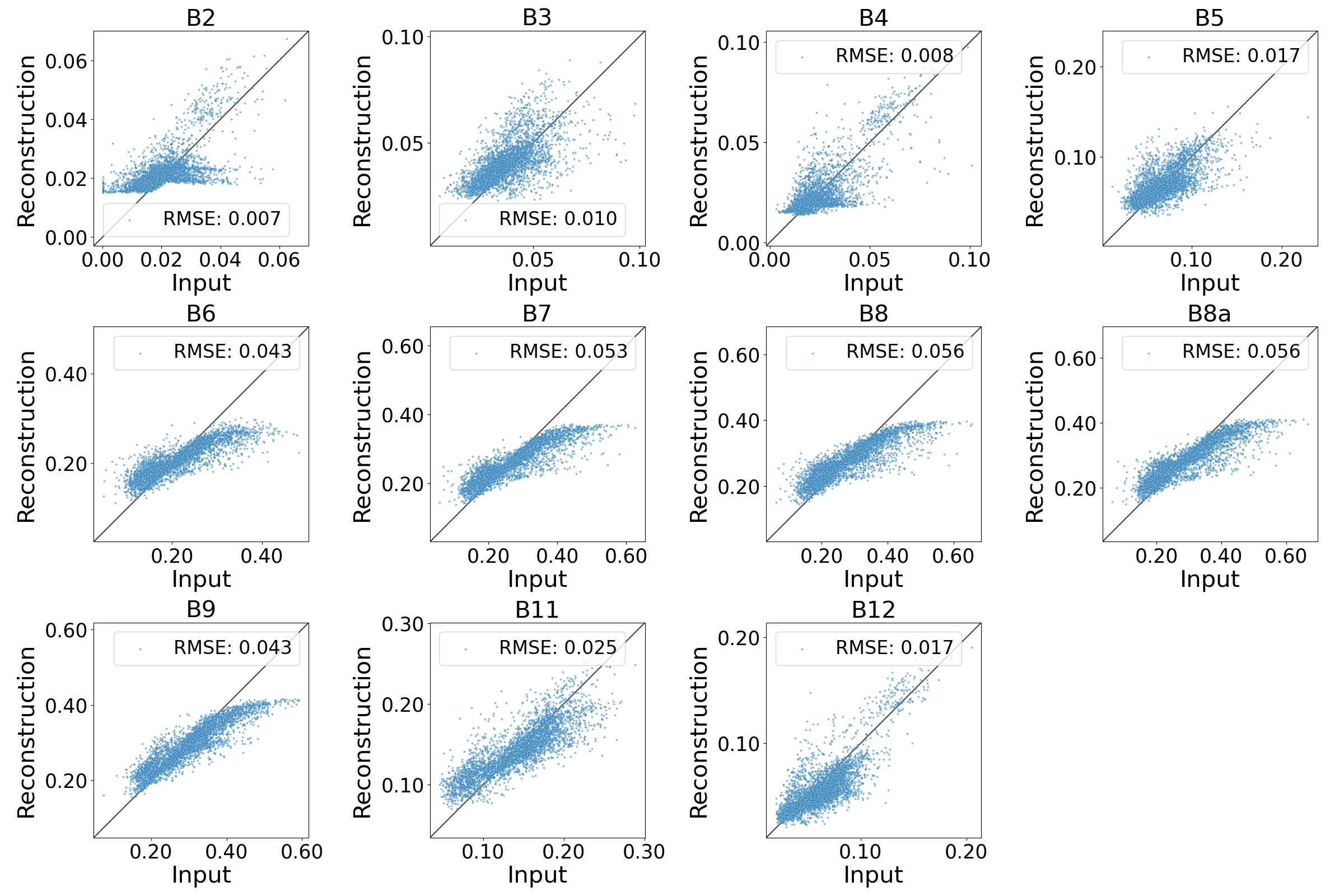}
        \caption{NNRegressor: $X_{r, D}$ v. $X_r$}
        \label{fig:linescatter_bands_nn}
    \end{subfigure}
    \begin{subfigure}{\textwidth}
        \centering
        \includegraphics[width=1\textwidth]{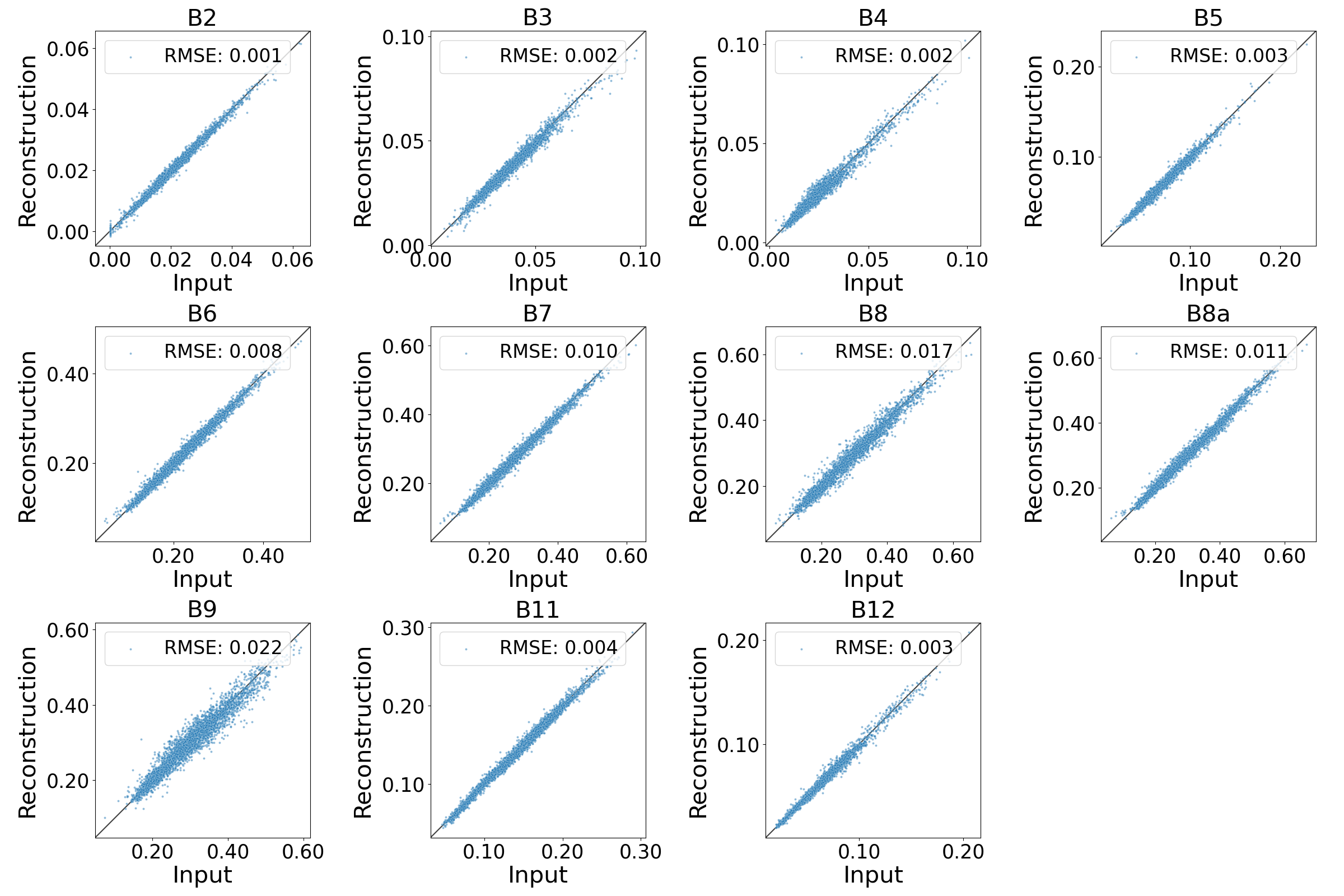}
        \caption{AE\_RTM\_corr: $X_{r, C}$ v. $X_r$}
        \label{fig:linescattter_bands_wocorr}
    \end{subfigure}
    
    \caption{\textbf{Superior reconstruction accuracy for AE\_RTM\_corr}, illustrated by spectral band. \textit{(a) Reconstruction from NNRegressor displays clear bias (b)  AE\_RTM\_corr  reconstructs $X_r$ markedly more accurately.}}
    \label{fig:linescatter_bands_both_nn_wocorr}
\end{figure}

The systematic bias of INFORM can be observed by comparing $X_r$ with $X_{r, D}$ based on the $Z_{r, D}$ that is inferred by NNRegressor (\cref{tab:models_mse_loss}). The distributions of the learned biases display distinct patterns among forest types (\cref{fig:hist_bias_corr_coni_v_deci}). The RTM tends to under-estimate the spectral components for deciduous forest but over-estimate for coniferous forest. Per-band analysis (\cref{fig:linescatter_bands_both_nn_wocorr}) indicates overestimation in the visible bands at lower ranges and underestimation in the near-infrared bands at higher ranges, with less pronounced bias in the shortwave bands.
Reconstruction loss with AE\_RTM\_corr is more than an order of magnitude lower than with NNRegressor (\cref{tab:models_mse_loss}). 
%
\begin{table*}[htbp] 
\caption{\textbf{MSE loss of trained models evaluated on $D_r$.} \textit{The Mean Squared Error (MSE) of AE\_RTM\_corr is more than an order of magnitude lower than NNRegressor. 
}}
\label{tab:models_mse_loss}
\centering
\begin{tabular}{cccccc}
\toprule
Model & Architecture & Dataset & $MSE_{train}$ & $MSE_{val}$ & $MSE_{test}$ \\
\midrule
AE\_RTM\_corr & $\mathbf{E}_{C}$+$\mathbf{F}$+$\mathbf{C}_{C}$ & $D_r$ & 0.0210 & 0.0235 &  0.0217 \\
NNRgressor & $\mathbf{E}_{D}$ & $D_r$ & - & - &  0.6676 \\
\bottomrule
\end{tabular}
\end{table*}

\subsection{Biophysical variables}\label{sec:vars}
\begin{figure}[htbp]
    \centering
    \begin{subfigure}{\textwidth}
        \centering
        \includegraphics[width=1\textwidth]{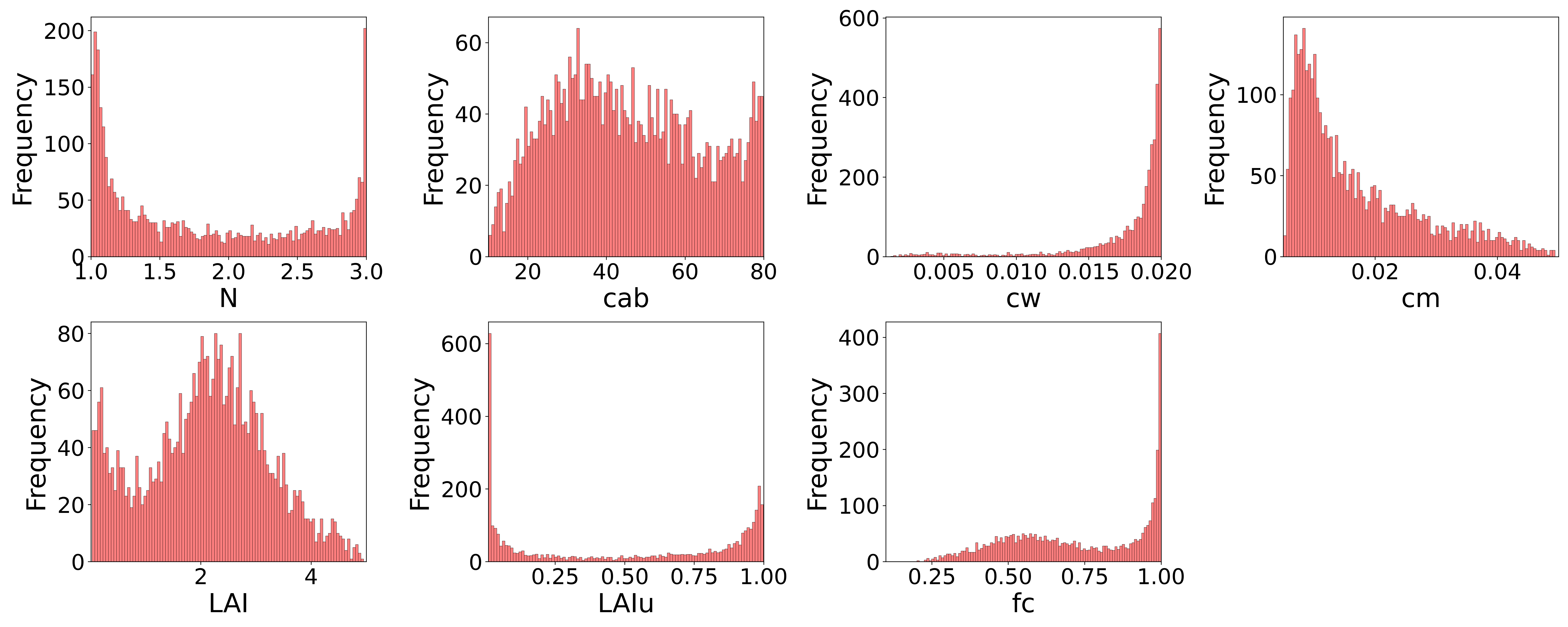}
        \caption{NNRegressor: $Z_{r, D}$}
        \label{fig:hist_vars_corr_v_nn}
    \end{subfigure}
    \begin{subfigure}{\textwidth}
        \centering
        \includegraphics[width=1\textwidth]{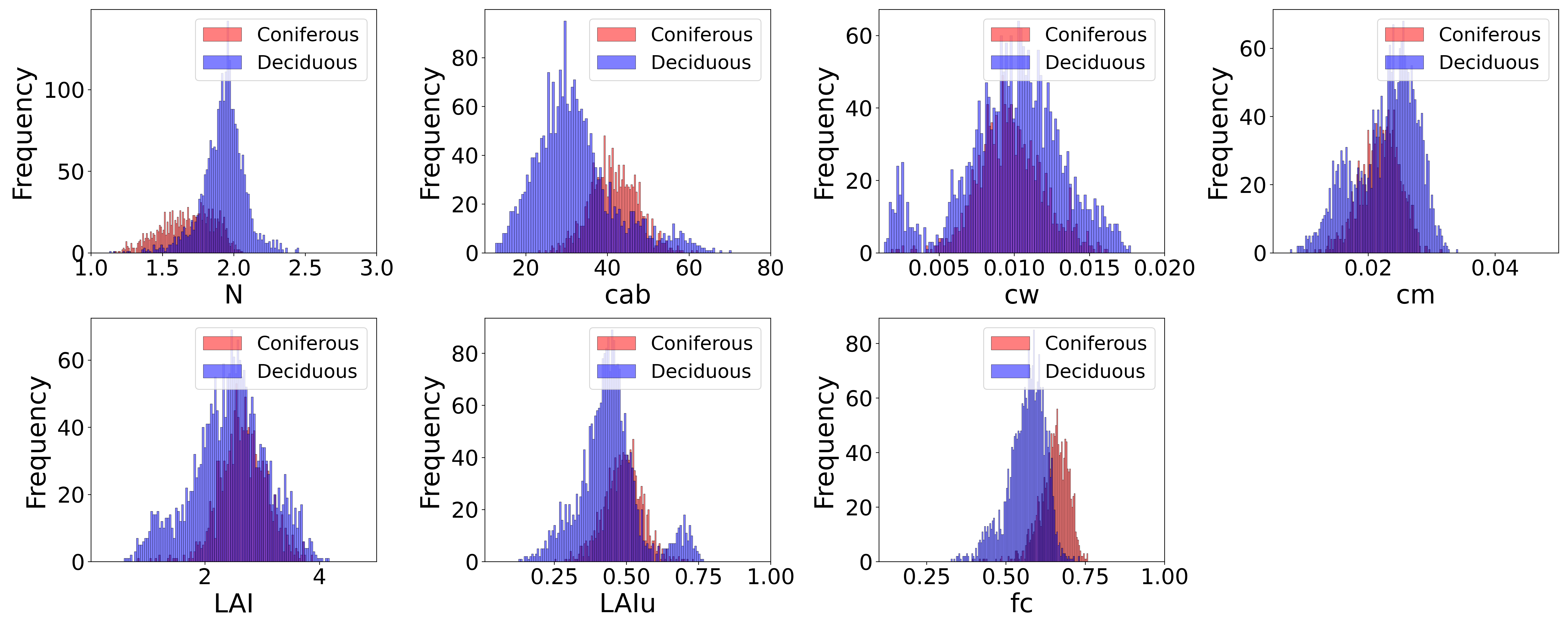}
        \caption{AE\_RTM\_corr: $Z_{r, C}$}
        \label{fig:hist_vars_corr_coni_v_deci}
    \end{subfigure}
    
    \caption{\textbf{Distributions of variables}. \textit{(a) Application of the NNRegressor to real Sentinel-2 spectra leads to implausible parameter distributions that tend to break out of the preset parameter ranges. (b) Distributions of variables learned by AE\_RTM\_corr are plausible and distinguish between forest types.}}
    \label{fig:hist_vars_corr_v_nn_wocorr}
\end{figure}

\begin{figure}[htbp]
    \centering
    \includegraphics[width=\textwidth]{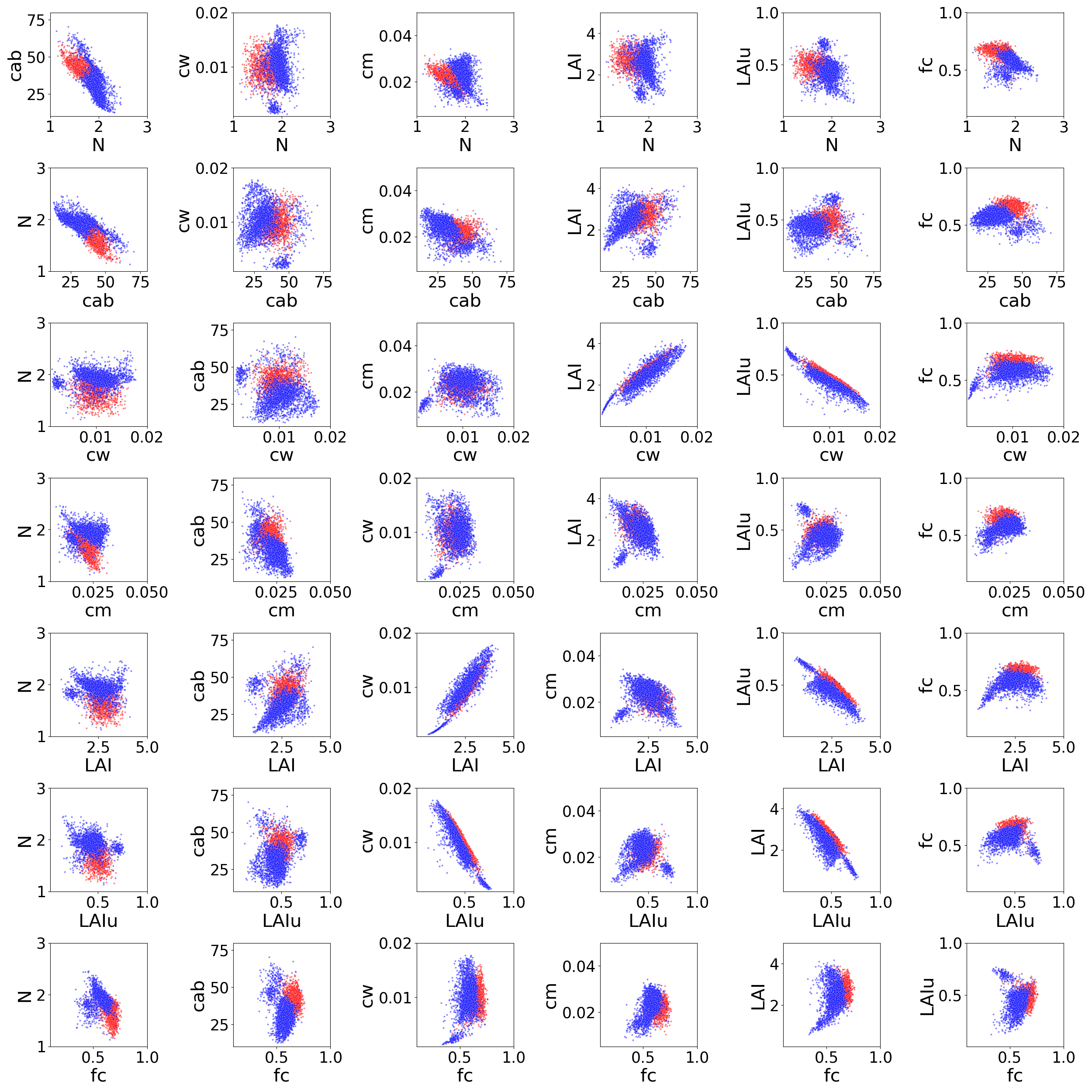}
    \caption{\textbf{Pairwise co-distributions of $Z_{r, C}$ learned by AE\_RTM\_corr}. \textit{Red: coniferous forest. Blue: deciduous forest. Our model can learn distinct physical patterns.}}
    \label{fig:scatter_vars_corr_coni_v_deci}
\end{figure}

The variables retrieved by our model follow a unimodal distribution (\cref{fig:hist_vars_corr_coni_v_deci}), whereas NNRegressor distorts the distribution of most variables to boost extreme values (\cref{fig:hist_vars_corr_v_nn}), tending to break out of the preset parameter ranges. 
We clustered the variables inferred by our model, by forest type (\cref{fig:hist_vars_corr_coni_v_deci}). Coniferous and deciduous forests have distinct distributions aligned with their respective attributes. 
%
Examining the co-distributions of variables, $LAI$ displays apparent positive correlation with $cw$ and negative correlation with $LAIu$ under AE\_RTM\_corr (\cref{fig:scatter_vars_corr_coni_v_deci}), while the variables learned by NNRegressor do not retain these correlation patterns. This makes sense as higher $LAI$ reduces light penetration causing lower plant growth in the understory, hence lower $LAIu$. 

\begin{figure}[htbp]
    \centering
    \includegraphics[width=\textwidth]{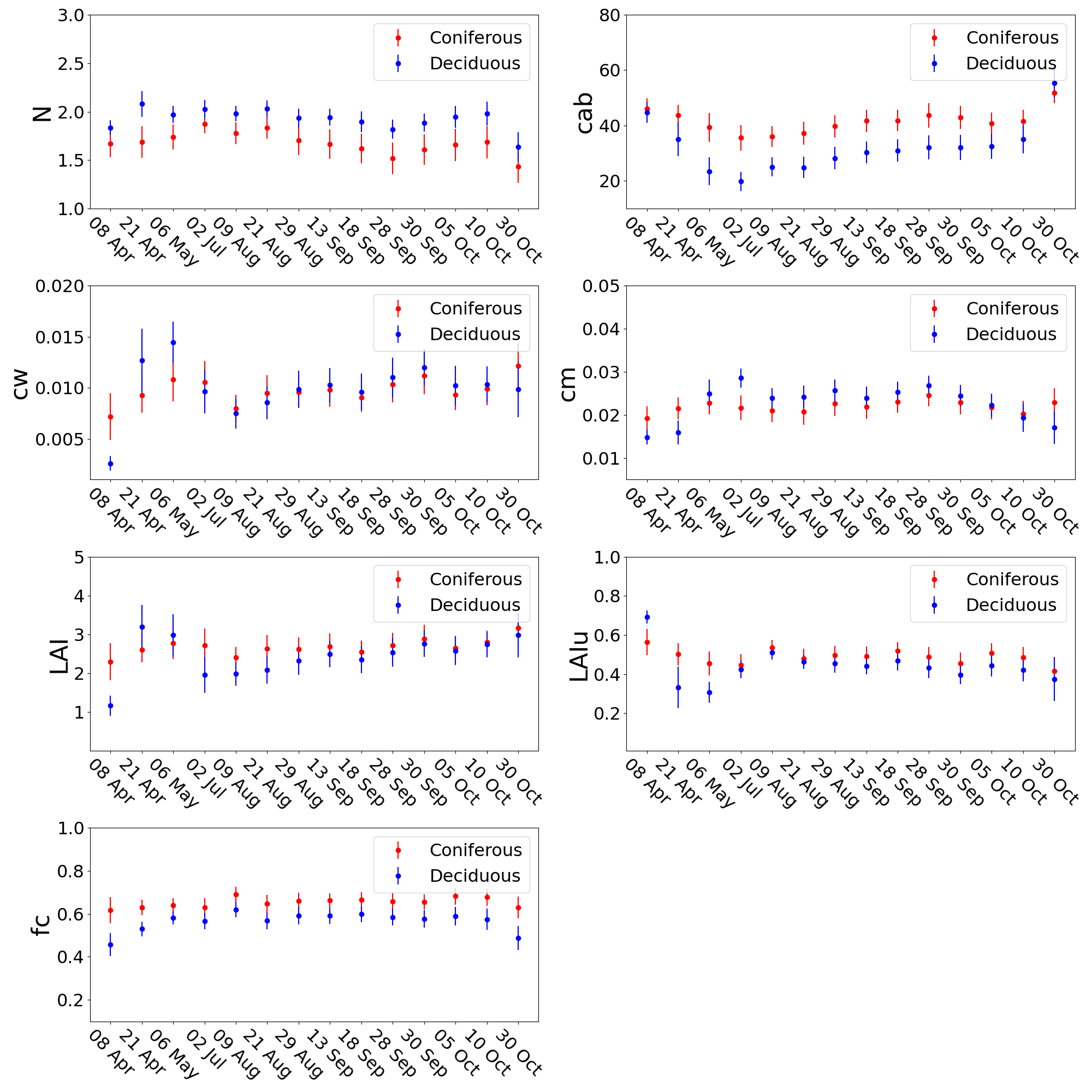}
    \caption{\textbf{Temporal variations of inferred physical parameters $Z_{r, C}$}. \textit{AE\_RTM\_corr effectively captures distinct, temporally smooth and plausible variations for different forest types.}}
    \label{fig:timeseries_vars_corr_coni_v_deci}
\end{figure}

Next, comparing temporal variations (\cref{fig:timeseries_vars_corr_coni_v_deci}), our model AE\_RTM\_corr shows more consistent and clearer patterns for some variables (\cref{fig:timeseries_vars_corr_coni_v_deci}). For example, understory plant growth ($LAIu$) declines from April due to reduced light penetration from increasing canopy growth ($LAI$); coniferous forests exhibit different patterns than deciduous forests, with higher fractional coverage ($fc$) and less variation over time, as would be expected. NNRegressor temporal variations appear to be less consistent (see \cref{fig:timeseries_vars_nn_coni_v_deci_appx}).

\begin{figure}[htbp]
    \centering
    \begin{subfigure}{0.5\textwidth}
        \centering
        \includegraphics[width=1\textwidth]{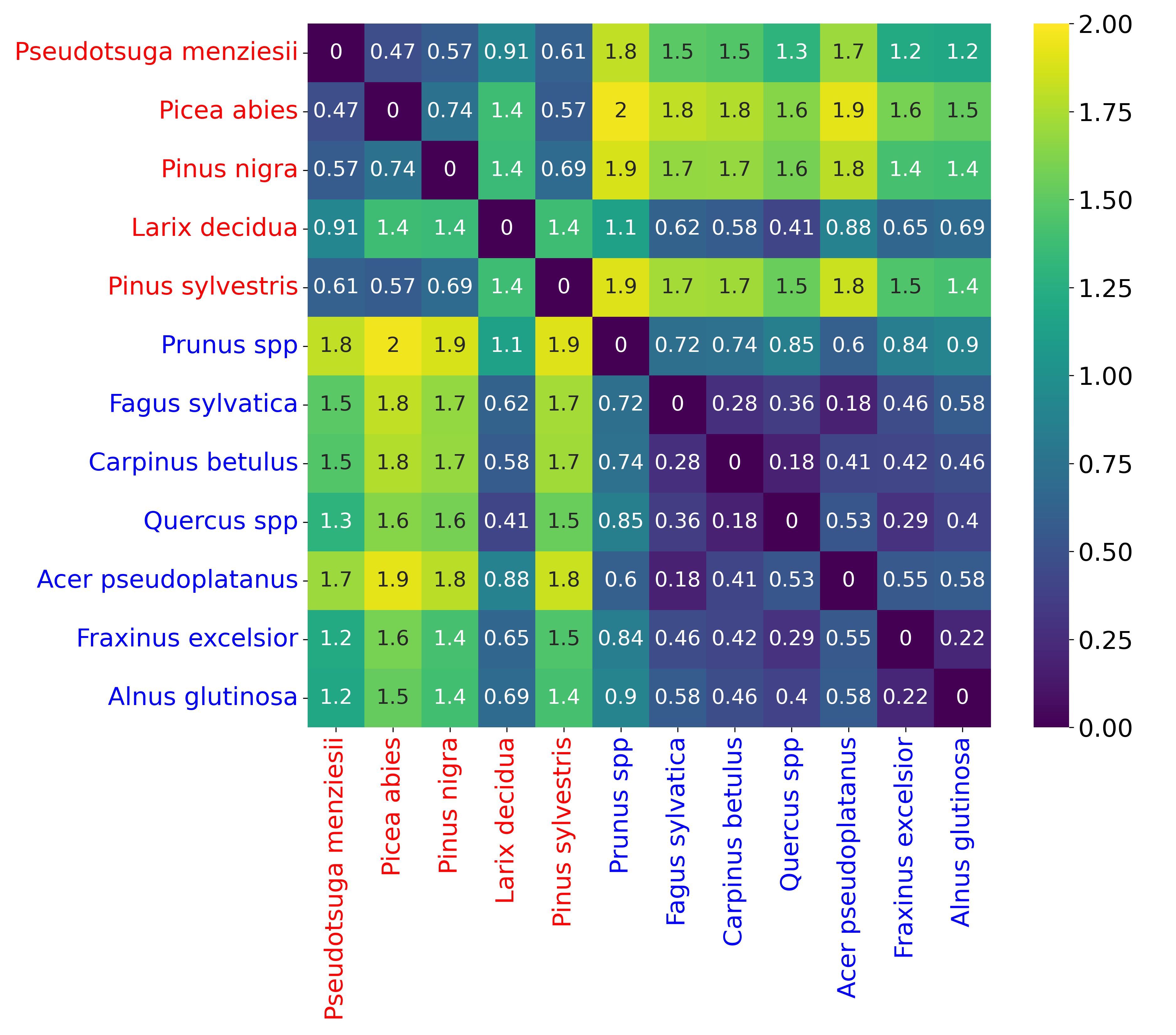}
        \caption{NNRegressor}
        \label{fig:heatmap_jm_dist_species_nn}
    \end{subfigure}\hfill
    \begin{subfigure}{0.5\textwidth}
        \centering
        \includegraphics[width=1\textwidth]{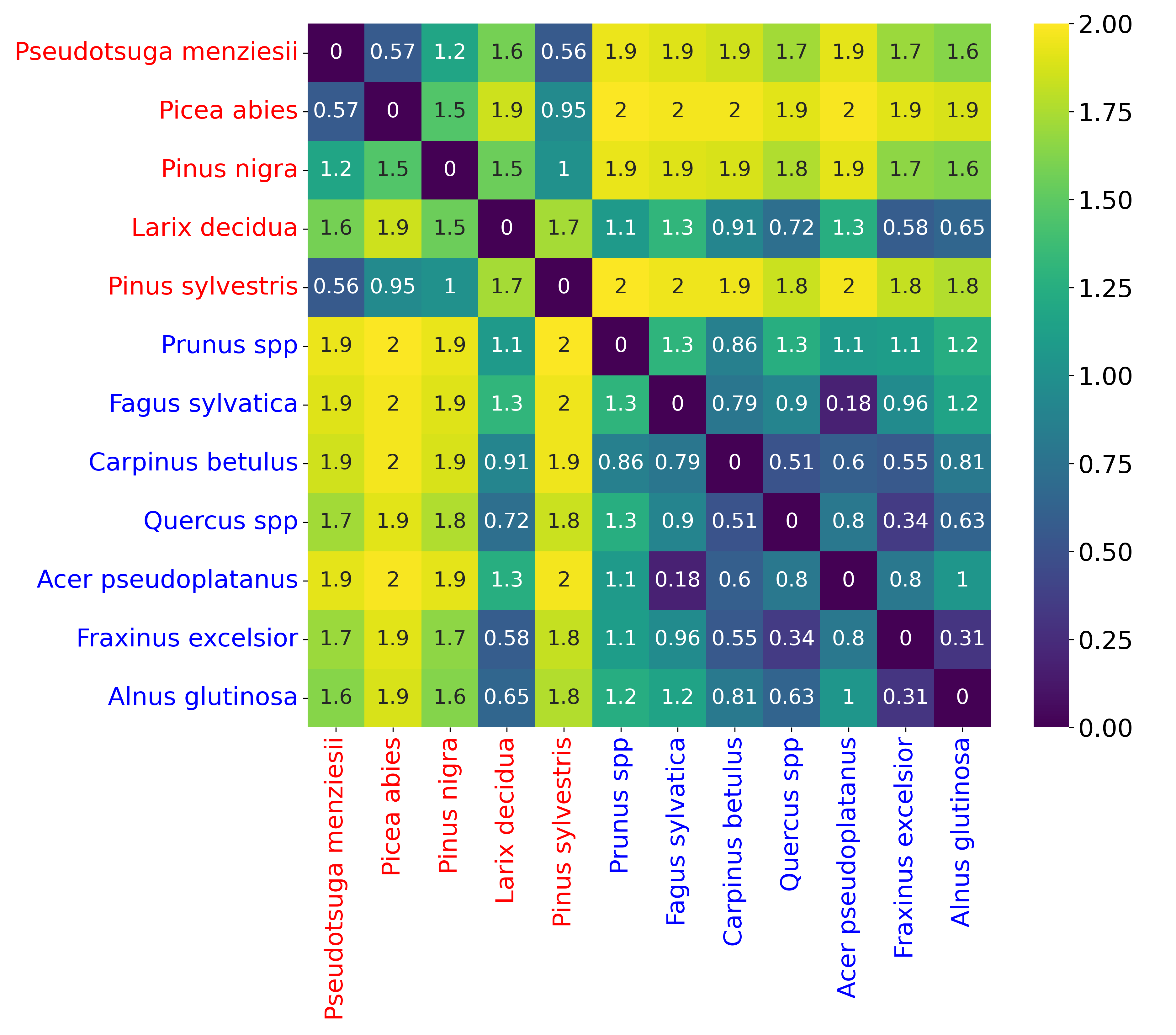}
        \caption{AE\_RTM\_corr}
        \label{fig:heatmap_jm_dist_species_corr}
    \end{subfigure}
    
    \caption{\textbf{Pairwise Jeffreys-Matusita (JM) distance between species} based on the learned variables. \textit{Red: coniferous species. Blue: deciduous species.  JM distance is a statistical measure used to quantify the separability between two probability distributions. Our model is able to learn more disentangled variables within the latent space.}}
    \label{fig:heatmap_jm_distance_nn_corr}
\end{figure}

We have clustered the 7 inferred variables by species and calculated the Jeffreys-Matusita (JM) distance to assess the separability of the 12 species based on these variables (refer to \cref{tab:stats_vars_nn_appx} and \cref{tab:stats_vars_corr_appx} for detailed statistics). The JM distance, which varies between 0 and 2, serves as an indicator of separability, with higher values denoting greater distinction between two distributions. Consequently, we computed the pairwise JM distances for the species represented by the seven biophysical variables. This quantifies the effectiveness of our model in differentiating the species. 
As shown in \cref{fig:heatmap_jm_distance_nn_corr}, our model achieves clear separation between species of different forest types and shows improved separability for species within the same forest type, indicating that the model is capable of inferring variables with clearer distinguishing power within the latent space.

\FloatBarrier
\subsection{Ablation studies}
To evaluate the effects of our experimental design, we have conducted ablation studies (\cref{tab:ablation_loss}). The systematic bias of the RTM can be observed by comparing the MSE of the classical autoencoder ($\mathbf{E}_{A}$+$\mathbf{D}_{A}$) with that of one where the decoder has been replaced by the RTM ($\mathbf{E}_{B}$+$\mathbf{F}$). In this comparison, the reconstruction accuracy becomes significantly worse—more than an order of magnitude—when the decoder is replaced by the RTM. However, with bias correction, our model achieves MSE losses comparable to those of the classical encoder.

\begin{figure}[h]
    \centering
    \includegraphics[width=1\textwidth]{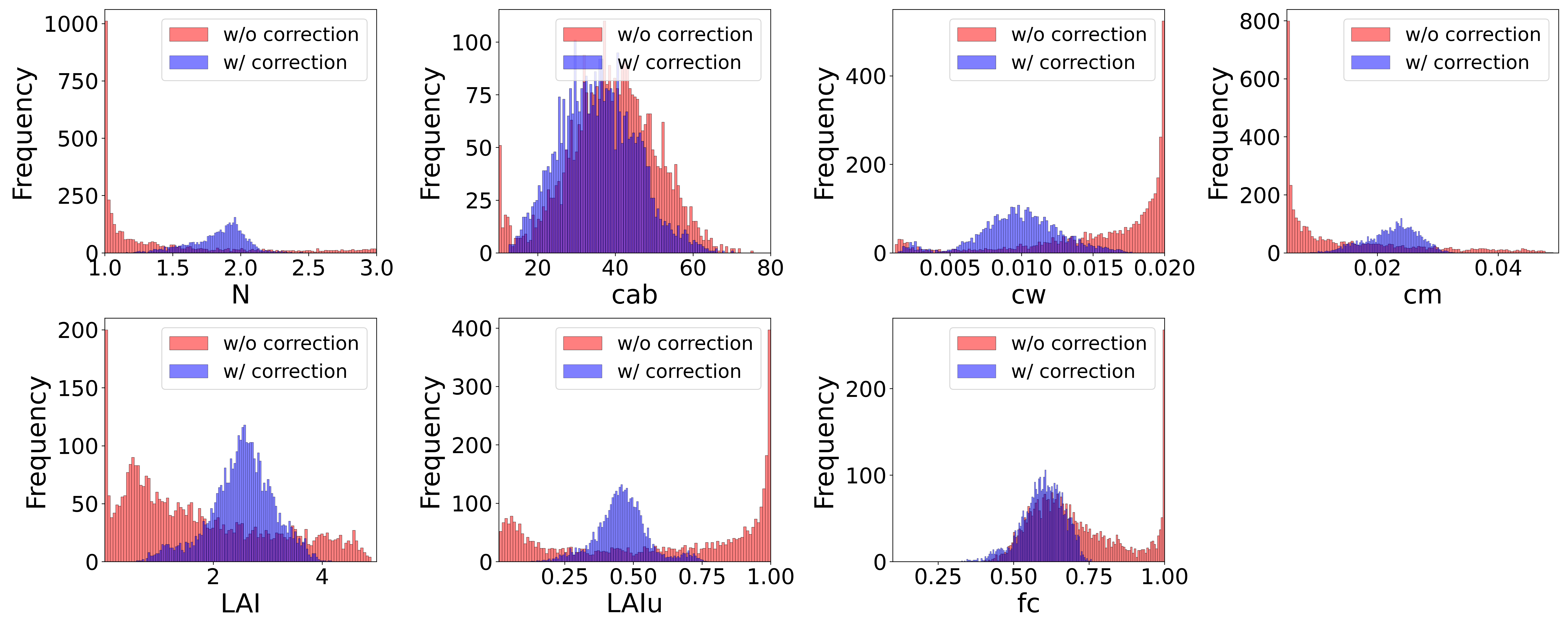}
    \caption{\textbf{Distributions of variables} from our model before and after bias correction ($Z_{r, B}$ v. $Z_{r, C}$). \textit{Without bias correction the variable distributions (red) are implausible, tending to break out beyond the preset bounded ranges. With bias correction (blue), the model is able to learn more plausible distributions.}}
    \label{fig:hist_vars_corr_v_wocorr}
\end{figure}

The effects of bias correction extend beyond improving reconstruction accuracy; they also enhance the learning of variables in the latent space. Comparing the retrieved variables of our model with and without the bias correction layer (\cref{fig:hist_vars_corr_v_wocorr}), it is evident that the RTM's bias distorts the distribution of several variables to extremes, likely in an attempt to minimize the reconstruction loss despite the presence of bias. In contrast, with bias correction, our model learns more plausible distributions of the variables.

\begin{table*}[htbp] 
\caption{\textbf{MSE loss of models under different steps of the inversion}. \textit{The MSE loss of our final model is more than one magnitude lower than the one without bias correction, and is comparable to the classical auto-encoder.}}
\label{tab:ablation_loss}
\centering
\begin{tabular}{cccccc}
\toprule
Model & Ablation & Dataset & $MSE_{train}$ & $MSE_{val}$ & $MSE_{test}$ \\
\midrule
$\mathbf{E}_{A}$+$\mathbf{D}_{A}$ & w/o RTM, w/o correction & $D_r$ & 0.0193 & 0.0219 & 0.0191 \\
$\mathbf{E}_{B}$+$\mathbf{F}$ & w/ RTM, w/o correction & $D_r$ & 0.0875 & 0.0833 & 0.0856 \\
$\mathbf{E}_{C}$+$\mathbf{F}$+$\mathbf{C}_{C}$ (\textbf{ours}) & w/ RTM, w/ correction & $D_r$ & 0.0210 & 0.0235 &  0.0217 \\
\bottomrule
\end{tabular}
\end{table*}
\FloatBarrier

\section{Conclusion}

We have developed an end-to-end method for extracting biophysical variables from satellite spectra. We integrated a complex radiative transfer model with modern machine learning, using auto-encoder learning to invert a physical model and correct its biases.  Our extensive analysis shows markedly improved reconstruction loss, and strongly suggests that recovered physical variables are more plausible than previous methods. There may also be broader implications for methodology in physics-informed machine learning.

\textbf{Acknowledgement:} This work was supported by the UKRI Centre for Doctoral Training in Application of Artificial Intelligence to the study of Environmental Risks (reference EP/S022961/1) and Cambridge Centre for Carbon Credits. We would also like to thank Markus Immitzer from Mantle Labs for having shared the Sentinel-2 data with us. 

\clearpage
\bibliography{iclr2024_conference}

\begin{thebibliography}{28}
\providecommand{\natexlab}[1]{#1}
\providecommand{\url}[1]{\texttt{#1}}
\expandafter\ifx\csname urlstyle\endcsname\relax
  \providecommand{\doi}[1]{doi: #1}\else
  \providecommand{\doi}{doi: \begingroup \urlstyle{rm}\Url}\fi

\bibitem[Atzberger(2000)]{atzberger2000development}
Clement Atzberger.
\newblock Development of an invertible forest reflectance model: The infor-model.
\newblock In \emph{A decade of trans-european remote sensing cooperation. Proceedings of the 20th EARSeL Symposium Dresden, Germany}, volume~14, pp.\  39--44, 2000.

\bibitem[Chen et~al.(2016)Chen, Duan, Houthooft, Schulman, Sutskever, and Abbeel]{chen2016infogan}
Xi~Chen, Yan Duan, Rein Houthooft, John Schulman, Ilya Sutskever, and Pieter Abbeel.
\newblock Infogan: Interpretable representation learning by information maximizing generative adversarial nets.
\newblock \emph{Advances in neural information processing systems}, 29, 2016.

\bibitem[Claverie et~al.(2018)Claverie, Ju, Masek, Dungan, Vermote, Roger, Skakun, and Justice]{claverie2018harmonized}
Martin Claverie, Junchang Ju, Jeffrey~G Masek, Jennifer~L Dungan, Eric~F Vermote, Jean-Claude Roger, Sergii~V Skakun, and Christopher Justice.
\newblock The harmonized landsat and sentinel-2 surface reflectance data set.
\newblock \emph{Remote sensing of environment}, 219:\penalty0 145--161, 2018.

\bibitem[Combal et~al.(2003)Combal, Baret, Weiss, Trubuil, Mace, Pragnere, Myneni, Knyazikhin, and Wang]{combal2003retrieval}
B~Combal, Fr{\'e}d{\'e}ric Baret, M~Weiss, Alain Trubuil, D~Mace, A~Pragnere, R~Myneni, Y~Knyazikhin, and L~Wang.
\newblock Retrieval of canopy biophysical variables from bidirectional reflectance: Using prior information to solve the ill-posed inverse problem.
\newblock \emph{Remote sensing of environment}, 84\penalty0 (1):\penalty0 1--15, 2003.

\bibitem[Cootes et~al.(1995)Cootes, Taylor, Cooper, and Graham]{cootes1995active}
Timothy~F Cootes, Christopher~J Taylor, David~H Cooper, and Jim Graham.
\newblock Active shape models-their training and application.
\newblock \emph{Computer vision and image understanding}, 61\penalty0 (1):\penalty0 38--59, 1995.

\bibitem[Eastwood \& Williams(2018)Eastwood and Williams]{eastwood2018framework}
Cian Eastwood and Christopher~KI Williams.
\newblock A framework for the quantitative evaluation of disentangled representations.
\newblock In \emph{International Conference on Learning Representations}, 2018.

\bibitem[Gastellu-Etchegorry et~al.(1996)Gastellu-Etchegorry, Demarez, Pinel, and Zagolski]{gastellu1996modeling}
Jean-Philippe Gastellu-Etchegorry, Val{\'e}rie Demarez, Veronique Pinel, and Francis Zagolski.
\newblock Modeling radiative transfer in heterogeneous 3-d vegetation canopies.
\newblock \emph{Remote sensing of environment}, 58\penalty0 (2):\penalty0 131--156, 1996.

\bibitem[Goel(1988)]{goel1988models}
Narendra~S Goel.
\newblock Models of vegetation canopy reflectance and their use in estimation of biophysical parameters from reflectance data.
\newblock \emph{Remote sensing reviews}, 4\penalty0 (1):\penalty0 1--212, 1988.

\bibitem[Gong(1999)]{gong1999inverting}
P~Gong.
\newblock Inverting a canopy reflectance model using a neural network.
\newblock \emph{International Journal of Remote Sensing}, 20\penalty0 (1):\penalty0 111--122, 1999.

\bibitem[Goodfellow et~al.(2020)Goodfellow, Pouget-Abadie, Mirza, Xu, Warde-Farley, Ozair, Courville, and Bengio]{goodfellow2020generative}
Ian Goodfellow, Jean Pouget-Abadie, Mehdi Mirza, Bing Xu, David Warde-Farley, Sherjil Ozair, Aaron Courville, and Yoshua Bengio.
\newblock Generative adversarial networks.
\newblock \emph{Communications of the ACM}, 63\penalty0 (11):\penalty0 139--144, 2020.

\bibitem[Hao et~al.(2022)Hao, Liu, Zhang, Ying, Feng, Su, and Zhu]{hao2022physics}
Zhongkai Hao, Songming Liu, Yichi Zhang, Chengyang Ying, Yao Feng, Hang Su, and Jun Zhu.
\newblock Physics-informed machine learning: A survey on problems, methods and applications.
\newblock \emph{arXiv preprint arXiv:2211.08064}, 2022.

\bibitem[Higgins et~al.(2017)Higgins, Matthey, Pal, Burgess, Glorot, Botvinick, Mohamed, and Lerchner]{higgins2017beta}
Irina Higgins, Loic Matthey, Arka Pal, Christopher Burgess, Xavier Glorot, Matthew Botvinick, Shakir Mohamed, and Alexander Lerchner.
\newblock beta-vae: Learning basic visual concepts with a constrained variational framework.
\newblock In \emph{International conference on learning representations}, 2017.

\bibitem[Jucker et~al.(2017)Jucker, Caspersen, Chave, Antin, Barbier, Bongers, Dalponte, van Ewijk, Forrester, Haeni, et~al.]{jucker2017allometric}
Tommaso Jucker, John Caspersen, J{\'e}r{\^o}me Chave, C{\'e}cile Antin, Nicolas Barbier, Frans Bongers, Michele Dalponte, Karin~Y van Ewijk, David~I Forrester, Matthias Haeni, et~al.
\newblock Allometric equations for integrating remote sensing imagery into forest monitoring programmes.
\newblock \emph{Global change biology}, 23\penalty0 (1):\penalty0 177--190, 2017.

\bibitem[Karras et~al.(2019)Karras, Laine, and Aila]{karras2019style}
Tero Karras, Samuli Laine, and Timo Aila.
\newblock A style-based generator architecture for generative adversarial networks.
\newblock In \emph{Proceedings of the IEEE/CVF conference on computer vision and pattern recognition}, pp.\  4401--4410, 2019.

\bibitem[Kingma \& Welling(2013)Kingma and Welling]{kingma2013auto}
Diederik~P Kingma and Max Welling.
\newblock Auto-encoding variational bayes.
\newblock \emph{arXiv preprint arXiv:1312.6114}, 2013.

\bibitem[Kumar et~al.(2018)Kumar, Sattigeri, and Balakrishnan]{kumar2018variational}
Abhishek Kumar, Prasanna Sattigeri, and Avinash Balakrishnan.
\newblock Variational inference of disentangled latent concepts from unlabeled observations, 2018.

\bibitem[Li \& Strahler(1985)Li and Strahler]{li1985geometric}
Xiaowen Li and Alan~H Strahler.
\newblock Geometric-optical modeling of a conifer forest canopy.
\newblock \emph{IEEE Transactions on Geoscience and Remote Sensing}, \penalty0 (5):\penalty0 705--721, 1985.

\bibitem[Locatello et~al.(2019)Locatello, Bauer, Lucic, Rätsch, Gelly, Schölkopf, and Bachem]{locatello2019challenging}
Francesco Locatello, Stefan Bauer, Mario Lucic, Gunnar Rätsch, Sylvain Gelly, Bernhard Schölkopf, and Olivier Bachem.
\newblock Challenging common assumptions in the unsupervised learning of disentangled representations, 2019.

\bibitem[Loper \& Black(2014)Loper and Black]{loper2014opendr}
Matthew~M Loper and Michael~J Black.
\newblock Opendr: An approximate differentiable renderer.
\newblock In \emph{Computer Vision--ECCV 2014: 13th European Conference, Zurich, Switzerland, September 6-12, 2014, Proceedings, Part VII 13}, pp.\  154--169. Springer, 2014.

\bibitem[Lu et~al.(2020)Lu, Kim, and Solja{\v{c}}i{\'c}]{lu2020extracting}
Peter~Y Lu, Samuel Kim, and Marin Solja{\v{c}}i{\'c}.
\newblock Extracting interpretable physical parameters from spatiotemporal systems using unsupervised learning.
\newblock \emph{Physical Review X}, 10\penalty0 (3):\penalty0 031056, 2020.

\bibitem[OpenAI(2023)]{openai2023gpt4}
OpenAI.
\newblock Gpt-4 technical report, 2023.

\bibitem[Rosema et~al.(1992)Rosema, Verhoef, Noorbergen, and Borgesius]{rosema1992new}
A~Rosema, W~Verhoef, H~Noorbergen, and JJ~Borgesius.
\newblock A new forest light interaction model in support of forest monitoring.
\newblock \emph{Remote Sensing of Environment}, 42\penalty0 (1):\penalty0 23--41, 1992.

\bibitem[Schlerf \& Atzberger(2006)Schlerf and Atzberger]{schlerf2006inversion}
Martin Schlerf and Clement Atzberger.
\newblock Inversion of a forest reflectance model to estimate structural canopy variables from hyperspectral remote sensing data.
\newblock \emph{Remote sensing of environment}, 100\penalty0 (3):\penalty0 281--294, 2006.

\bibitem[Suits(1971)]{suits1971calculation}
Gwynn~H Suits.
\newblock The calculation of the directional reflectance of a vegetative canopy.
\newblock \emph{Remote Sensing of Environment}, 2:\penalty0 117--125, 1971.

\bibitem[Thompson et~al.(2009)Thompson, Mackey, McNulty, Mosseler, et~al.]{thompson2009forest}
I~Thompson, B~Mackey, S~McNulty, A~Mosseler, et~al.
\newblock Forest resilience, biodiversity, and climate change.
\newblock In \emph{A synthesis of the biodiversity/resilience/stability relationship in forest ecosystems. Secretariat of the Convention on Biological Diversity, Montreal. Technical Series}, volume~43, pp.\  1--67, 2009.

\bibitem[Widlowski et~al.(2013)Widlowski, Pinty, Lopatka, Atzberger, Buzica, Chelle, Disney, Gastellu-Etchegorry, Gerboles, Gobron, et~al.]{widlowski2013fourth}
J-L Widlowski, B~Pinty, M~Lopatka, C~Atzberger, D~Buzica, Micha{\"e}l Chelle, M~Disney, J-P Gastellu-Etchegorry, M~Gerboles, N~Gobron, et~al.
\newblock The fourth radiation transfer model intercomparison (rami-iv): Proficiency testing of canopy reflectance models with iso-13528.
\newblock \emph{Journal of Geophysical Research: Atmospheres}, 118\penalty0 (13):\penalty0 6869--6890, 2013.

\bibitem[Z{\'e}rah et~al.(2022)Z{\'e}rah, Valero, and Inglada]{zerah2022physics}
Yo{\"e}l Z{\'e}rah, Silvia Valero, and Jordi Inglada.
\newblock Physics-guided interpretable probabilistic representation learning for high resolution image time series.
\newblock \emph{IEEE Transactions on Geoscience and Remote Sensing}, 2022.

\bibitem[Z{\'e}rah et~al.(2023)Z{\'e}rah, Valero, and Inglada]{zerah2023physics}
Yo{\"e}l Z{\'e}rah, Silvia Valero, and Jordi Inglada.
\newblock Physics-constrained deep learning for biophysical parameter retrieval from sentinel-2 images: inversion of the prosail model.
\newblock \emph{Available at SSRN 4671923}, 2023.

\end{thebibliography}
\bibliographystyle{iclr2024_conference}

\newpage
\appendix
\section{Appendix}
\setcounter{figure}{0}
\setcounter{table}{0}
\renewcommand{\thefigure}{A.\arabic{figure}}
\renewcommand{\thetable}{A.\arabic{table}}

\subsection{Conversion of INFORM to PyTorch assisted by GPT-4}\label{appx:rtm_conversion}
The original modules of INFORM are extracted from a vegetation app called EnMap \footnote{\href{https://enmap-box-lmu-vegetation-apps.readthedocs.io/en/latest/tutorials/IVVRM_tut.html}{EnMap}}. It is implemented using Numpy arrays and operations. However, to track the computational graph and enable gradient backpropagation, it needs to be reimplemented using PyTorch operations. However, INFORM is a complicated physical model, and reimplementing it in PyTorch manually seems challenging. In light of the recent development of large language models, we decide to utilize GPT-4 \citep{openai2023gpt4} to assist in the conversion from NumPy to PyTorch. While GPT-4 has helped us cut out a significant amount of repetitive work, it is worth mentioning a few of its limitations:

GPT-4 is useful for converting commonly seen operations from NumPy into PyTorch. With its assistance, we have successfully converted 1,742 lines of code across various scripts from the original implementation to PyTorch (e.g. Listing \ref{lst:np} and Listing \ref{lst:torch}). However, some operations e.g. the exponential integral function \footnote{\href{https://docs.scipy.org/doc/scipy/reference/generated/scipy.special.exp1.html}{scipy.special.exp1}}, has no equivalent in PyTorch (Listing \ref{lst:torch}). To calculate their derivative, we will still need to implement our own \texttt{backward} functions to define derivatives.

Additionally, GPT-4 only accepts text input of limited length, which means it can only convert a script snippet piece by piece. As the conversation lengthens, it may make mistakes. For instance, it converts \texttt{numpy.radians} to \texttt{torch.radians}, which does not exist in PyTorch. Thus it is still important to visually check the conversion, run unit tests, and work on compiling all the scripts together. Interestingly, GPT allows for prompt engineering, which means we can provide feedback to refine its conversions, such as requesting it to highlight uncertain parts for review in subsequent conversions.

\begin{lstlisting}[language=Python, caption={\textbf{Example code of $\mathbf{F}$ initially implemented using NumPy.} \textit{It features an exponential integral function 'exp1' in line 9. Additionally, it's important to highlight that in line 3, the structure parameter $N$ functions as a denominator in the calculation, therefore, it cannot be zero.}}, label=lst:np]
n = PD_refractive
k = (np.outer(Cab, PD_k_Cab) + np.outer(Car, PD_k_Car) + np.outer(Anth, PD_k_Anth) +
     np.outer(Cbrown, PD_k_Brown) + np.outer(Cw, PD_k_Cw) + np.outer(Cm, PD_k_Cm)) / N[:, np.newaxis]

ind_k0_row, ind_k0_col = np.where(k == 0)  

if len(ind_k0_row) > 0:
    k[ind_k0_row, ind_k0_col] = np.finfo(float).eps
trans = (1 - k) * np.exp(-k) + (k ** 2) * exp1(k)
\end{lstlisting}

\begin{lstlisting}[language=Python, caption={\textbf{Conversion of $\mathbf{F}$ to PyTorch by GPT.} \textit{It is followed by feedback on exp1 to raise user's attention. Here is an excerpt: "Please note that the exp1 function is not available in PyTorch as of my last training cut-off in September 2021. We can import it from the scipy.special library, but we have to convert the tensor to numpy before using it and convert the result back to a tensor. It's not ideal in terms of performance, but I'm afraid there's no other option for now. Please check if there is an updated version of PyTorch that includes this function."}}, label=lst:torch]
n = PD_refractive
k = (torch.outer(Cab, PD_k_Cab) + torch.outer(Car, PD_k_Car) + torch.outer(Anth, PD_k_Anth) +
     torch.outer(Cbrown, PD_k_Brown) + torch.outer(Cw, PD_k_Cw) + torch.outer(Cm, PD_k_Cm)) / N.unsqueeze(-1)

ind_k0_row, ind_k0_col = torch.where(k == 0)

if len(ind_k0_row) > 0:
    k[ind_k0_row, ind_k0_col] = torch.finfo(float).eps
trans = (1 - k) * torch.exp(-k) + (k ** 2) * torch.from_numpy(exp1(k.numpy()))
\end{lstlisting}

Note that this experiment was done in May, 2023, when GPT-4 was still in its early phase and has several features updated since then. For instance, it now can accept longer text inputs. Nonetheless, we believe the limitations discussed above still largely hold true and are worth considering for similar tasks in the future. 

\subsection{Validating the PyTorch implementation of INFORM}
To validate the implementation of INFORM in PyTorch against its original version in Numpy, before generating $D_s$, we randomly sample 10,000 sets of variables, passed them through each implementation, and compare their outputs (\cref{tab:conversion_unittest}). The mismatch rate of the outputs is 0.457\% when the absolute tolerance is set to 1e-5. The maximum absolute difference over all 130000 output reflectances from the two implementations is 3.050e-5 wherein the physical unit of reflectance here is 1. This indicates that INFORM implemented in PyTorch is substantially equivalent to its original version, therefore can be used in subsequent tasks. 
 
\begin{table*}[htbp] 
\caption{\textbf{Unit test of the outputs of our PyTorch implementation of $\mathbf{F}$ against the original NumPy version.} \textit{The total number of elements to compare is 130000 as we simulate the same set of 10000 samples for both implementations, each with 13 spectral bands of Sentinel-2.}}
\label{tab:conversion_unittest}
\centering
\resizebox{\textwidth}{!}{
    \begin{tabular}{ccccc}
    \toprule
    Total Elements & Mismatched Elements & Absolute Tolerance & Mismatch Ratio & Max Absolute Difference \\
    \midrule
    130000 &  594 &  1e-5 &  0.457\% &  3.050e-5 \\
    \bottomrule
    \end{tabular}
}
\end{table*}

\subsection{Integrating INFORM into machine learning}
\subsubsection{Numerical instability of INFORM}
With INFORM now implemented in PyTorch, we can backpropagate gradients through this physical model. However, the model is sensitive to input variable ranges and is non-differentiable at certain points. Furthermore, complex operations like exponential, logarithm, and square root, often lead to numerical instability during derivative computation, even though these operations are theoretically differentiable. To solve these issues, we have developed specific methods to stabilize the training when INFORM is used. 

\subsubsection{Forward pass} \label{appx:foward_pass}
The input variables $Z$ of INFORM are non-negative values with physical meanings (\cref{tab:para_list}). In practice, INFORM can fail in the forward pass when $Z$ from other value ranges are provided. Thus we choose to infer a normalised variable $\Lambda$ ranging from 0 to 1 for each biophysical variable; then we map $\Lambda$ to each biophysical variable in its original scale using the value ranges provided in \cref{tab:para_list} (\cref{eq:scaling_factor_to_z}). This not only enables a successful forward pass of INFORM but also narrows down the search space for optimization. 

\begin{equation}\label{eq:scaling_factor_to_z}
    Z = (Max - Min)\cdot\Lambda + Min
\end{equation}

\subsubsection{Backpropagation} \label{appx:backprop}
Despite a successful forward pass, the backpropagation, however, remains unstable. For successful learning, it is necessary to backpropagate gradients through the physical model i.e. INFORM in our case. The aforementioned caveats of INFORM however have resulted in backpropagation failures, leading to NaN gradient and then NaN loss. During experiments to stabilize the backpropagation, we notice that NaN gradients, when they initially appear, are always in association with specific sets of INFORM variables. Ideally, one would apply constraints to specific INFORM operations causing these NaN gradients. However, working out such constraints seems challenging unless we understand INFORM and its differentiability thoroughly.

In response to this, we introduce a simple workaround: since it is NaN gradients that cause the subsequent learning failure, we replace them with a small random value, which is sampled from a uniform distribution ranging from 0 to 1 and subsequently scaled by 1e-7, whenever these gradients first appear. The specific workflow of the algorithm is described in \cref{algo:grad_stab}

\begin{algorithm}
\caption{Gradient stabilization}\label{algo:grad_stab}
    \begin{algorithmic}
    \STATE Calculate gradients    
    \STATE Initialize list $grads$, containing gradients of all model parameters where gradients exist and contain any NaN values
    
    \IF{$grads$ is not empty}
        \STATE Set a small constant epsilon equal to 1e-7

        \FOR{each v in $grads$}
            \STATE Generate random values of the same shape as v, scaled by epsilon
            \STATE Create a mask where v is NaN or equals 0
            \STATE Replace values in v where the mask is True with corresponding random values
        \ENDFOR
    \ENDIF
    \STATE Update gradients    
    \end{algorithmic}
\end{algorithm}

\subsubsection{Stabilized training process} \label{appx:stab_train}
\begin{figure}[htbp]
    \centering
    \includegraphics[width=1\textwidth]{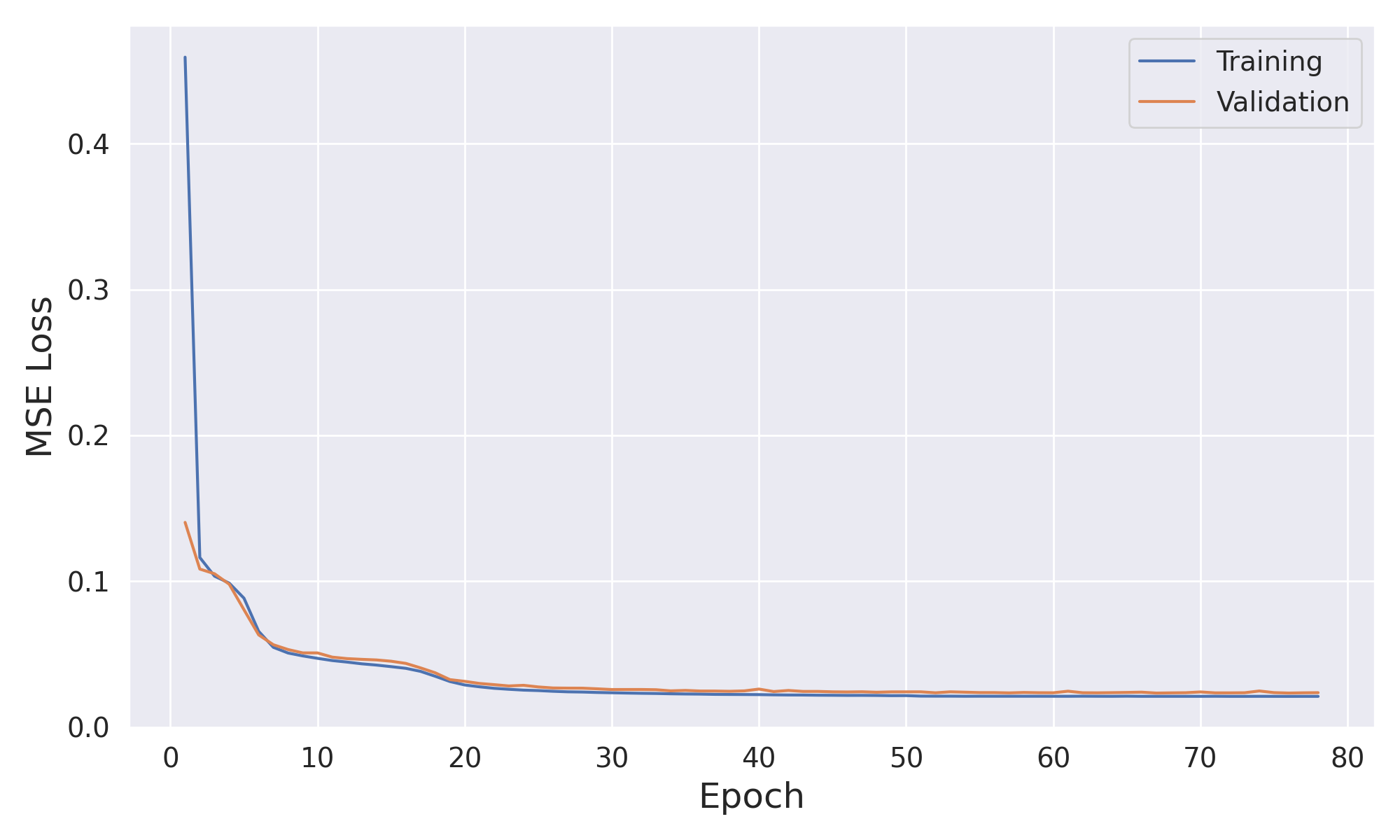}
    \caption{\textbf{Training and validation losses of AE\_RTM\_corr.} \textit{Our strategy to update the gradients has overcome the instability of $\mathbf{F}$ during backpropagation and allowed the convergence of training loss.}}
    \label{fig:ae_rtm_corr_train_val_loss}
\end{figure}

Our method has effectively stabilized the learning of $\mathbf{E}_B$ and $\mathbf{E}_C$ in the autoencoder framework, despite instability of $\mathbf{F}$ during backpropagation.  As can be seen in \cref{fig:ae_rtm_corr_train_val_loss}, both training and validation losses of AE\_RTM\_corr keep converging over epochs. Therefore, by preventing the propagation of NaN gradients to earlier layers and allowing the normal progression of the forward pass, the stabilizer has effectively enabled the training process to bypass instability points in the optimization space and allowed a continuous search for optimal points.

\subsection{Biophysical variables of INFORM}\label{appx:vars}
The input of INFORM consists of biophysical variables of three hierarchical levels. Note that the original INFORM does not have fractional coverage $fc$ as an input variable. We include $fc$ as one of the seven variables to be learned directly by $\mathbf{E}$, which will be used to infer crown diameter $cd$ and height $h$ based on derived equations. 
As the fractional coverage is jointly defined by the stem density and crown diameter within each unit hector (or 10,000 $m^2$), $cd$ can be derived given $fc$ and $cd$ using \cref{eq:fc2cd}. Furthermore, to derive $h$, we fit an allometric equation (\cref{eq:cd2h}) using the samples of temperate forests from the global allometric database \citep{jucker2017allometric}. $R^2$ of the derived equation between $h$ and $cd$ \cref{eq:cd2h} is 0.383.

\begin{figure}[h]
    \centering
    \includegraphics[width=1\textwidth]{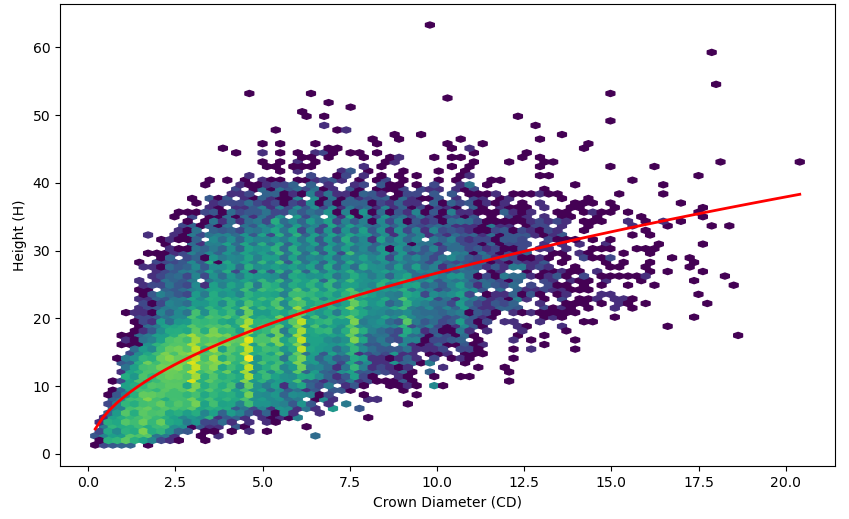}
    \caption{\textbf{Allometric relationship between $h$ and $cd$.} \textit{We fit an allometric equation between $h$ and $cd$ (\cref{eq:cd2h}). $R^2$ of the fitted equation is 0.383.}}
    \label{fig:ae_rtm_corr_train_val_loss_appx}
\end{figure}

\begin{equation}\label{eq:fc2cd}
    cd = 2 \cdot \sqrt{\frac{fc \cdot 10000}{\pi \cdot sd}}
\end{equation}

\begin{equation}\label{eq:cd2h}
    h = \exp\left(2.117 + 0.507 \cdot \ln(cd)\right)
\end{equation}

\subsection{Temporal and species information of the real dataset}
Statistics of the $D_r$ can be viewed in \cref{tab:real_data_appx}. These spectra were sampled from invidual sites covering a time sequence of 14 timestamps. Note that our current training strategy does not integrate temporal information into the inferencing process, although temporal variations are evaluated. Such temporal information could serve as useful prior knowledge to boost the model's performance, especially to ensure the consistency of temporal variations, which we will consider in future works. All these samples cover both coniferous and deciduous forests consisting of 5 and 7 species, respectively:
\begin{itemize}
    \item \textbf{Coniferous forest}: Pseudotsuga Menziesii, Picea Abies, Pinus Nigra, Larix Decidua, Pinus Sylvestris.
    \item \textbf{Deciduous forest}: Prunus Spp, Fagus Sylvatica, Carpinus Betulus', Quercus Spp, Acer Pseudoplatanus, Fraxinus Excelsior', Alnus Glutinosa.
\end{itemize}

\begin{table*}
\caption{Statistics of the real dataset}
\label{tab:real_data_appx}
\centering
\resizebox{\textwidth}{!}{
    \begin{tabular}{cccc}
    \toprule
    Total Number of Spectra & Number of Individual Sites & Number of Dates & Number of Species\\
    \midrule
    17962 & 1283 &  14 &  12 \\
    \bottomrule
    \end{tabular}
}
\end{table*}

\FloatBarrier
\subsection{Bias correction}
\begin{figure}[h]
    \centering
    \includegraphics[width=0.95\textwidth]{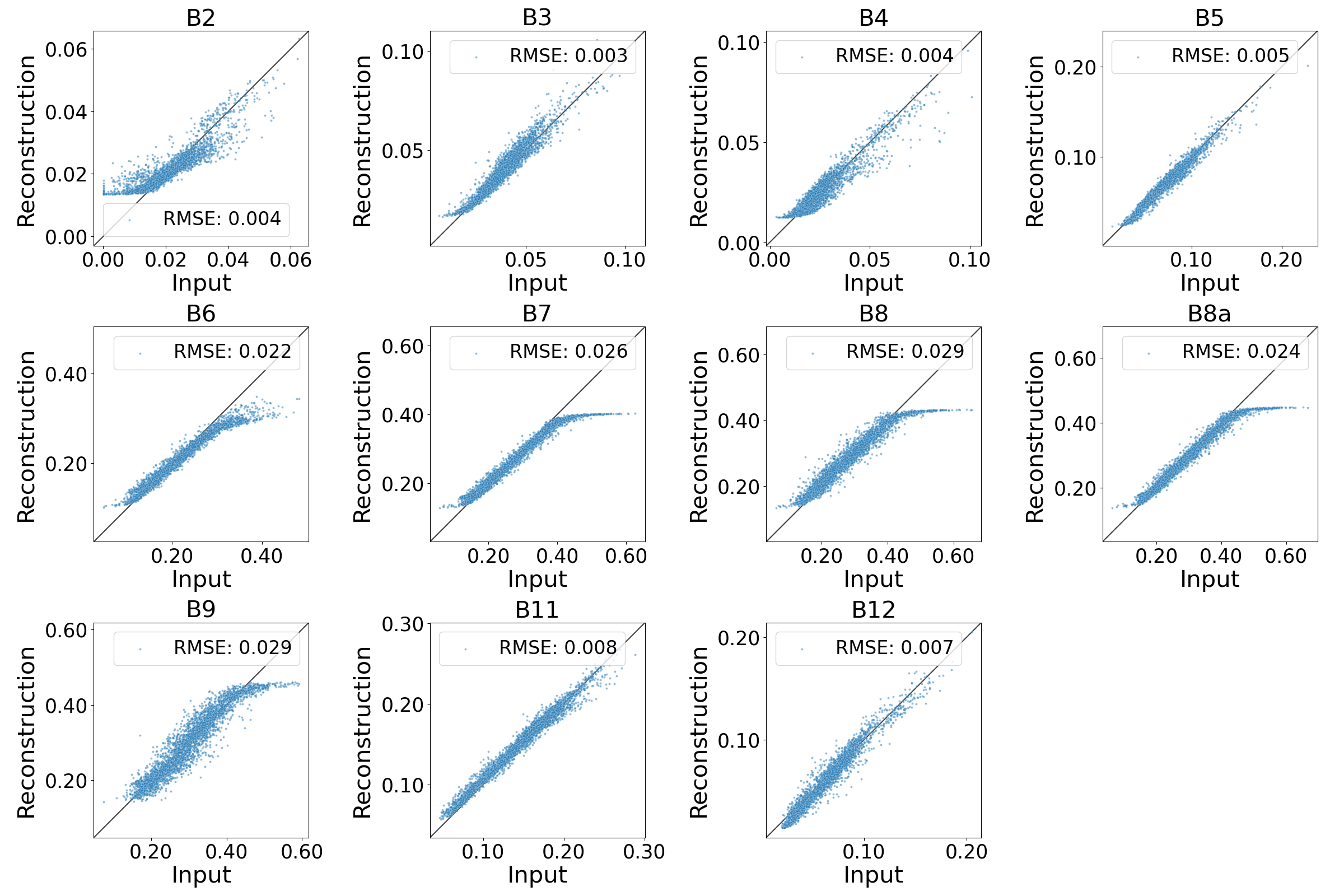}
    \caption{\textbf{Reconstruction visualization of our model without bias correction ($X_r$ against $X_{r, B}$)}. \textit{Without the correction by $\mathbf{C}_C$, $X_{r, B}$ also shows bias with similar patterns as NNRegressor.}}
    \label{fig:linescatter_bands_wocorr_appx}
\end{figure}

\begin{figure}[h]
    \centering
    \includegraphics[width=1\textwidth]{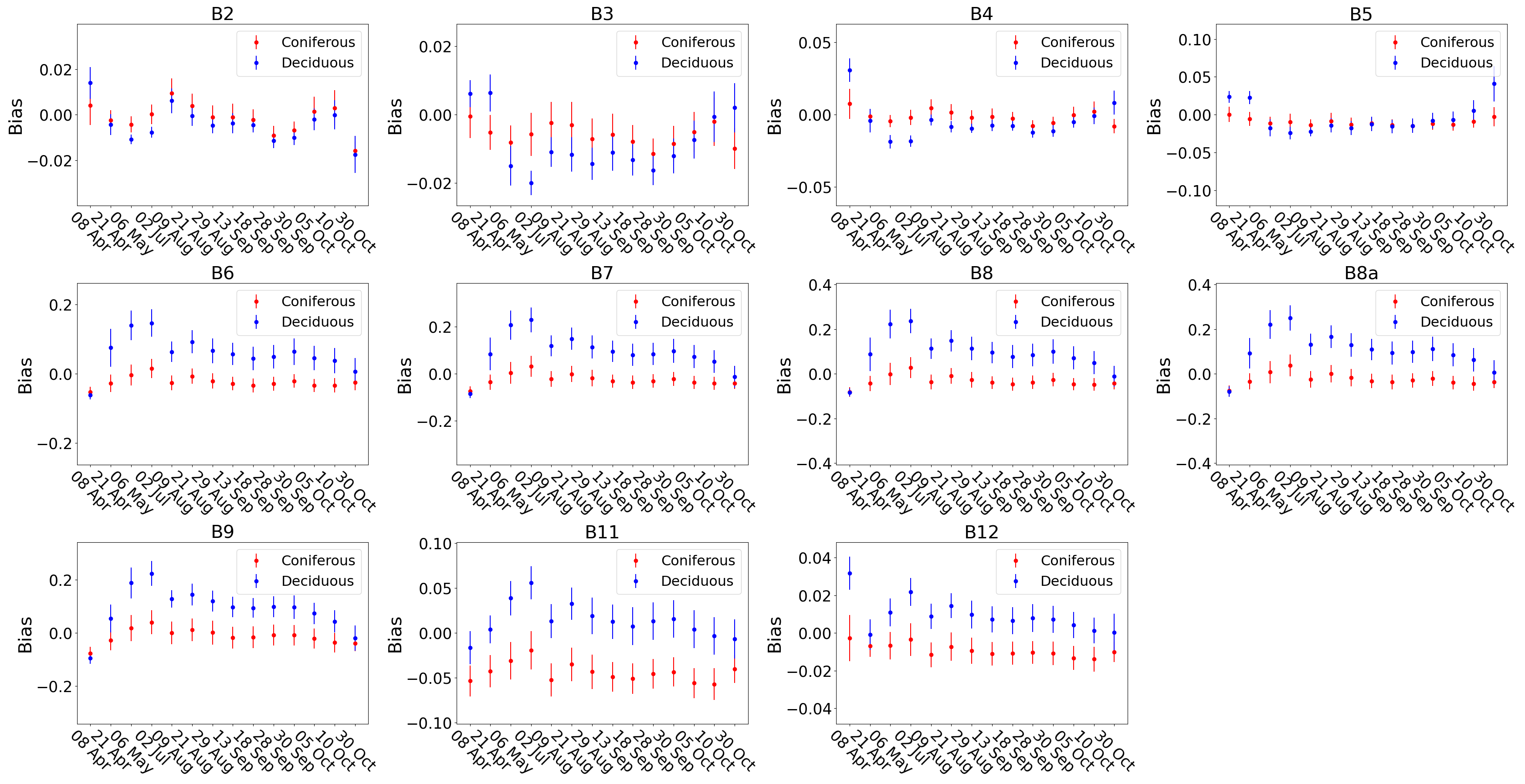}
    \caption{\textbf{Temporal variations of biases learned by AE\_RTM\_corr}. \textit{Biases from the same spectral group, e.g. those in the near-infrared, display similar patterns, which are also distinct between forest types.}}
    \label{fig:timeseries_bias_corr_coni_v_deci_appx}
\end{figure}

\begin{figure}[h]
    \centering
    \begin{subfigure}{0.5\textwidth}
        \centering
        \includegraphics[width=1\textwidth]{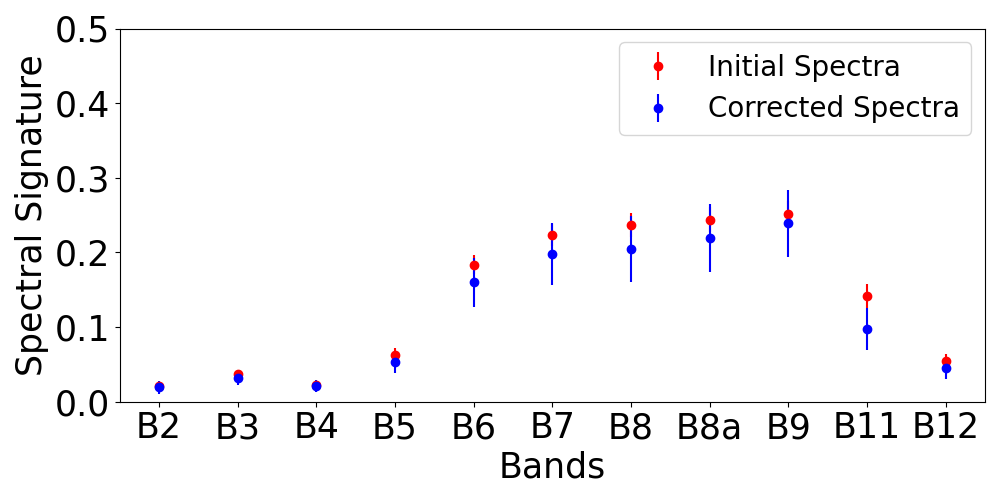}
        \caption{Coniferous forest}
        \label{fig:spec_sig_corr_init_v_final_coni_appx}
    \end{subfigure}\hfill
    \begin{subfigure}{0.5\textwidth}
        \centering
        \includegraphics[width=1\textwidth]{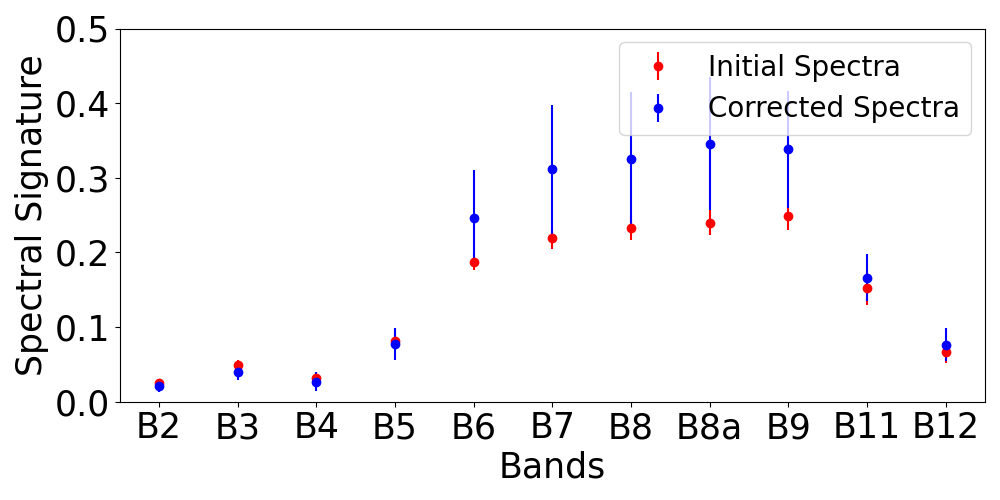}
        \caption{Deciduous forest}
        \label{fig:spec_sig_corr_init_v_final_deci_appx}
    \end{subfigure}
    
    \caption{\textbf{Spectral signature} before and after bias correction. \textit{The RTM tends to over-estimate the spectra for coniferous forest and under-estimate for deciduous forest.}}
    \label{fig:signature_corr_coni_v_deci_appx}
\end{figure}

Without bias correction, our model struggles to accurately reconstruct the input bands (\cref{fig:linescatter_bands_wocorr_appx}), displaying reconstruction patterns akin to the baseline. Specifically, for the visible bands, the model tends to overestimate values at lower ranges. In the case of near-infrared bands, it tends to underestimate at higher ranges, whereas the bias in the shortwave band is less pronounced.
The biases identified by our model (\cref{fig:timeseries_bias_corr_coni_v_deci_appx}) exhibit temporally consistent and smooth patterns, mirroring the temporal variations observed in the variables.
As revealed by the histogram of biases (\cref{fig:hist_bias_corr_coni_v_deci}), the spectral signatures simulated by the original RTM and our AE\_RTM\_corr, for the same set of retrieved variables, also reflect the tendency of the RTM to overestimate for coniferous and underestimate for deciduous species, respectively (\cref{fig:signature_corr_coni_v_deci_appx}).

\FloatBarrier
\subsection{Biophysical variables}
\begin{figure}[htbp]
    \centering
    \includegraphics[width=1\textwidth]{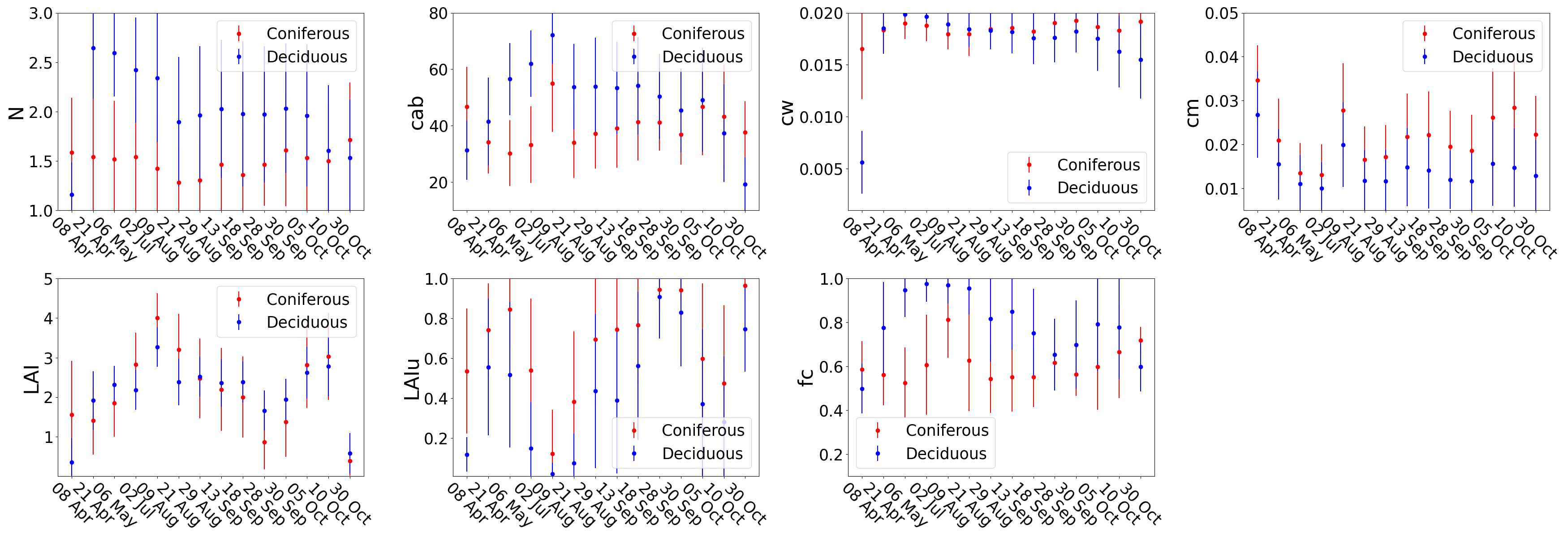}
    \caption{\textbf{Temporal variation of the $Z_{r, D}$} learned by NNRegressor, clustered by forest types. \textit{Compared to our model, the temporal variations of the variables retrieved by the baseline model are less consistent over time.}}
     \label{fig:timeseries_vars_nn_coni_v_deci_appx}
\end{figure}

The baseline model's learning of temporal variations in variables demonstrates increased variability over time, often reaching extreme values at the boundaries of specified ranges (\cref{fig:timeseries_vars_nn_coni_v_deci_appx}). Additionally, we have presented tables summarizing the means and standard deviations of the learned variables, categorized by species as identified by the baseline (\cref{tab:stats_vars_nn_appx}) and by our model (\cref{tab:stats_vars_corr_appx}). Our analysis, particularly of the Jeffreys-Matusita distance of species distributions based on learned variables (\cref{fig:heatmap_jm_distance_nn_corr}), indicates that our model succeeds in learning variables with more distinguishing power within the latent space.

\begin{table*}[htbp] 
\caption{\textbf{Statistics of $Z_{r, D}$ learned by NNRegressor}}
\label{tab:stats_vars_nn_appx}
\centering
\resizebox{\textwidth}{!}{
\begin{tabular}{lccccccc}
\toprule
Species & N & cab & cw & cm & LAI & LAIu & fc \\ 
\midrule
Pseudotsuga menziesii & 1.43 $\pm$ 0.54 & 38.31 $\pm$ 14.87 & 1.79 $\pm$ 0.25 & 1.91 $\pm$ 1.02 & 2.15 $\pm$ 1.24 & 0.72 $\pm$ 0.34 & 0.61 $\pm$ 0.19 \\ 
Picea abies & 1.36 $\pm$ 0.44 & 39.49 $\pm$ 13.24 & 1.87 $\pm$ 0.17 & 2.15 $\pm$ 0.98 & 1.72 $\pm$ 1.25 & 0.84 $\pm$ 0.25 & 0.58 $\pm$ 0.13 \\ 
Pinus nigra & 1.56 $\pm$ 0.55 & 39.50 $\pm$ 15.81 & 1.86 $\pm$ 0.20 & 2.47 $\pm$ 1.13 & 2.32 $\pm$ 1.42 & 0.54 $\pm$ 0.37 & 0.62 $\pm$ 0.19 \\ 
Larix decidua & 1.50 $\pm$ 0.50 & 38.26 $\pm$ 13.85 & 1.61 $\pm$ 0.42 & 1.43 $\pm$ 0.69 & 1.90 $\pm$ 1.04 & 0.56 $\pm$ 0.40 & 0.67 $\pm$ 0.21 \\ 
Pinus sylvestris & 1.56 $\pm$ 0.58 & 42.34 $\pm$ 15.68 & 1.90 $\pm$ 0.11 & 1.99 $\pm$ 1.01 & 2.40 $\pm$ 1.25 & 0.67 $\pm$ 0.37 & 0.59 $\pm$ 0.18 \\ 
Prunus spp & 1.90 $\pm$ 0.84 & 43.22 $\pm$ 22.17 & 1.35 $\pm$ 0.57 & 1.60 $\pm$ 1.04 & 1.79 $\pm$ 0.82 & 0.46 $\pm$ 0.34 & 0.71 $\pm$ 0.25 \\ 
Fagus sylvatica & 1.98 $\pm$ 0.76 & 46.75 $\pm$ 20.22 & 1.71 $\pm$ 0.42 & 1.21 $\pm$ 0.79 & 2.17 $\pm$ 0.92 & 0.33 $\pm$ 0.37 & 0.82 $\pm$ 0.23 \\ 
Carpinus betulus & 1.96 $\pm$ 0.71 & 47.55 $\pm$ 19.40 & 1.67 $\pm$ 0.36 & 1.36 $\pm$ 0.84 & 1.91 $\pm$ 0.91 & 0.55 $\pm$ 0.40 & 0.75 $\pm$ 0.21 \\ 
Quercus spp & 1.91 $\pm$ 0.69 & 48.59 $\pm$ 18.28 & 1.73 $\pm$ 0.41 & 1.46 $\pm$ 0.83 & 2.09 $\pm$ 0.94 & 0.55 $\pm$ 0.40 & 0.75 $\pm$ 0.21 \\ 
Acer pseudoplatanus & 2.21 $\pm$ 0.75 & 51.34 $\pm$ 20.31 & 1.72 $\pm$ 0.45 & 1.51 $\pm$ 0.79 & 2.01 $\pm$ 0.99 & 0.27 $\pm$ 0.33 & 0.86 $\pm$ 0.18 \\ 
Fraxinus excelsior & 2.16 $\pm$ 0.71 & 51.67 $\pm$ 18.05 & 1.76 $\pm$ 0.37 & 1.93 $\pm$ 1.17 & 2.05 $\pm$ 1.07 & 0.44 $\pm$ 0.39 & 0.78 $\pm$ 0.21 \\ 
Alnus glutinosa & 2.22 $\pm$ 0.72 & 55.01 $\pm$ 18.09 & 1.78 $\pm$ 0.34 & 2.03 $\pm$ 0.97 & 2.04 $\pm$ 1.03 & 0.46 $\pm$ 0.41 & 0.76 $\pm$ 0.20 \\ 

\bottomrule
\end{tabular}
}
\end{table*}

\begin{table*}[htbp] 
\caption{\textbf{Statistics of $Z_{r, C}$ learned by AE\_RTM\_corr}}
\label{tab:stats_vars_corr_appx}
\resizebox{\textwidth}{!}{
\centering
\begin{tabular}{lccccccc}
\toprule
Species & N & cab & cw & cm & LAI & LAIu & fc \\ 
\midrule
Pseudotsuga menziesii & 1.65 $\pm$ 0.16 & 39.96 $\pm$ 5.33 & 1.00 $\pm$ 0.24 & 2.36 $\pm$ 0.23 & 2.66 $\pm$ 0.43 & 0.49 $\pm$ 0.07 & 0.66 $\pm$ 0.04 \\ 
Picea abies & 1.53 $\pm$ 0.16 & 43.33 $\pm$ 4.89 & 0.99 $\pm$ 0.24 & 2.37 $\pm$ 0.21 & 2.70 $\pm$ 0.43 & 0.50 $\pm$ 0.07 & 0.68 $\pm$ 0.04 \\ 
Pinus nigra & 1.75 $\pm$ 0.16 & 42.71 $\pm$ 5.63 & 0.96 $\pm$ 0.19 & 1.97 $\pm$ 0.28 & 2.73 $\pm$ 0.37 & 0.48 $\pm$ 0.06 & 0.64 $\pm$ 0.04 \\ 
Larix decidua & 1.86 $\pm$ 0.13 & 36.52 $\pm$ 7.50 & 0.88 $\pm$ 0.25 & 2.28 $\pm$ 0.33 & 2.34 $\pm$ 0.48 & 0.49 $\pm$ 0.08 & 0.59 $\pm$ 0.05 \\ 
Pinus sylvestris & 1.71 $\pm$ 0.14 & 39.87 $\pm$ 5.18 & 0.99 $\pm$ 0.18 & 2.24 $\pm$ 0.24 & 2.68 $\pm$ 0.34 & 0.48 $\pm$ 0.05 & 0.66 $\pm$ 0.03 \\ 
Prunus spp & 1.93 $\pm$ 0.12 & 34.71 $\pm$ 8.65 & 0.80 $\pm$ 0.30 & 2.26 $\pm$ 0.37 & 2.13 $\pm$ 0.54 & 0.50 $\pm$ 0.10 & 0.53 $\pm$ 0.05 \\ 
Fagus sylvatica & 1.98 $\pm$ 0.17 & 29.76 $\pm$ 10.71 & 0.94 $\pm$ 0.33 & 2.24 $\pm$ 0.53 & 2.32 $\pm$ 0.69 & 0.43 $\pm$ 0.11 & 0.54 $\pm$ 0.06 \\ 
Carpinus betulus & 1.90 $\pm$ 0.11 & 33.38 $\pm$ 9.65 & 0.99 $\pm$ 0.31 & 2.37 $\pm$ 0.48 & 2.42 $\pm$ 0.61 & 0.45 $\pm$ 0.10 & 0.56 $\pm$ 0.05 \\ 
Quercus spp & 1.87 $\pm$ 0.11 & 33.46 $\pm$ 8.55 & 1.06 $\pm$ 0.33 & 2.35 $\pm$ 0.48 & 2.56 $\pm$ 0.63 & 0.44 $\pm$ 0.10 & 0.59 $\pm$ 0.06 \\ 
Acer pseudoplatanus & 1.99 $\pm$ 0.14 & 30.82 $\pm$ 10.09 & 0.99 $\pm$ 0.34 & 2.12 $\pm$ 0.49 & 2.45 $\pm$ 0.68 & 0.42 $\pm$ 0.12 & 0.54 $\pm$ 0.05 \\ 
Fraxinus excelsior & 1.90 $\pm$ 0.12 & 34.73 $\pm$ 8.27 & 1.02 $\pm$ 0.32 & 2.19 $\pm$ 0.39 & 2.57 $\pm$ 0.56 & 0.45 $\pm$ 0.10 & 0.58 $\pm$ 0.05 \\ 
Alnus glutinosa & 1.89 $\pm$ 0.13 & 35.02 $\pm$ 8.37 & 0.99 $\pm$ 0.27 & 2.22 $\pm$ 0.30 & 2.52 $\pm$ 0.48 & 0.46 $\pm$ 0.08 & 0.59 $\pm$ 0.05 \\ 

\bottomrule
\end{tabular}
}
\end{table*}

\end{document}